\documentclass[english]{IEEEtran}
\usepackage[T1]{fontenc}
\usepackage[latin9]{inputenc}
\usepackage{color}
\usepackage{verbatim}
\usepackage{float}
\usepackage{amsmath}
\usepackage{amsthm}
\usepackage{amssymb}
\usepackage{stmaryrd}
\usepackage{stackrel}
\usepackage{graphicx}
\usepackage{nicefrac}  
\usepackage{wasysym}
\usepackage{mathtools}
\usepackage{hyperref}
\usepackage{url}   
\usepackage{booktabs} 
\usepackage{nicefrac} 
\usepackage{subfig}
\usepackage[dvipsnames]{xcolor}
\makeatletter

\floatstyle{ruled}
\newfloat{algorithm}{tbp}{loa}
\providecommand{\algorithmname}{Algorithm}
\floatname{algorithm}{\protect\algorithmname}

\theoremstyle{plain}
\newtheorem{thm}{\protect\theoremname}
\theoremstyle{definition}
\newtheorem{defn}{\protect\definitionname}
\theoremstyle{definition}
\newtheorem{problem}{\protect\problemname}
\theoremstyle{plain}
\newtheorem{lem}{\protect\lemmaname}
\theoremstyle{definition}
\newtheorem{example}{\protect\examplename}
\theoremstyle{remark}
\newtheorem{rem}{\protect\remarkname}
\theoremstyle{plain}
\newtheorem{cor}{\protect\corollaryname}
\usepackage{tikz}
\tikzset{>=latex}
\usetikzlibrary{fit,automata,positioning,angles,quotes}

\usepackage{cite}
\usepackage{algpseudocode}
\usepackage{setspace}
\IEEEoverridecommandlockouts

\usepackage{fancyhdr}
\makeatother

\usepackage{babel}
\providecommand{\corollaryname}{Corollary}
\providecommand{\definitionname}{Definition}
\providecommand{\examplename}{Example}
\providecommand{\lemmaname}{Lemma}
\providecommand{\problemname}{Problem}
\providecommand{\remarkname}{Remark}
\providecommand{\theoremname}{Theorem}

\usepackage{amsmath}               
  {
      \theoremstyle{plain}
      
  }

\SetSymbolFont{stmry}{bold}{U}{stmry}{m}{n}

\def\BibTeX{{\rm B\kern-.05em{\sc i\kern-.025em b}\kern-.08em
		T\kern-.1667em\lower.7ex\hbox{E}\kern-.125emX}}
\begin{document}

\title{Receding Horizon Control Based Online Motion Planning with Partially
Infeasible LTL Specifications\thanks{$^{1}$Department of Mechanical Engineering, The University of Iowa,
Iowa City, IA, 52246, USA.}\thanks{$^{2}$Apex.AI Inc, Palo Alto, CA, 94303, USA.}\thanks{$^{3}$Department of Automation, University of Science and Technology
of China, Hefei, Anhui, China.}}
\author{Mingyu Cai$^{1}$, Hao Peng $^{2}$, Zhijun Li$^{3}$, Hongbo Gao$^{3}$,
and Zhen Kan $^{3}$}
\maketitle
\thispagestyle{fancy}

\begin{abstract}
This work considers online optimal motion planning of an autonomous
agent subject to linear temporal logic (LTL) constraints. The environment
is dynamic in the sense of containing mobile obstacles and time-varying
areas of interest (i.e., time-varying reward and workspace properties)
to be visited by the agent. Since user-specified tasks may not be
fully realized (i.e., partially infeasible), this work considers hard
and soft LTL constraints, where hard constraints enforce safety \textcolor{black}{requirements}
(e.g. avoid obstacles) while soft constraints represent tasks that
can be relaxed to not strictly follow user specifications. The motion
planning of the agent is to generate policies, in decreasing order
of priority, to 1) guarantee the satisfaction of safety constraints;
2) mostly satisfy\textcolor{black}{{} fulfill con}straints (i.e., minimize
the violation cost if desired tasks are partially infeasible); and
3) optimize the objective of rewards collection (i.e., visiting dynamic
areas of more interests). To achieve these objectives, a relaxed product
automaton, which allows the agent to not strictly follow the desired
LTL constraints, is constructed. A utility function is developed to
quantify the differences between the revised and the desired motion
plan, and the accumulated rewards are designed to bias the motion
plan towards those areas of more interests. Receding horizon control
is synthesized with an LTL formula to maximize the accumulated utilities
over a finite horizon, while ensuring that safety constraints are
fully satisfied and soft constraints are mostly satisfied. Simulation
and experiment results are provided to demonstrate the effectiveness
of the developed motion strategy.

\begin{IEEEkeywords}
Formal Method, Model Predictive Control, Multi-Objective Optimization, Graph Theory
\end{IEEEkeywords}

\global\long\def\Dist{\operatorname{Dist}}%
\global\long\def\Inf{\operatorname{Inf}}%
\global\long\def\Sense{\operatorname{Sense}}%
\global\long\def\Eval{\operatorname{Eval}}%
\global\long\def\Info{\operatorname{Info}}%
\global\long\def\ResetRabin{\operatorname{ResetRabin}}%
\global\long\def\ResetRabinState{\operatorname{ResetRabinState}}%
\global\long\def\Path{\operatorname{Path}}%
\global\long\def\Reward{\operatorname{Reward}}%
\end{abstract}

\section{Introduction}

Motion planning of autonomous agents has broad potential applications
ranging from driverless cars navigating urban environments subject
to complex traffic rules \cite{Luettel2012} to autonomous vehicles
(e.g., an unmanned ground or aerial vehicle) performing search and
rescue missions in uncertain environments after a natural disaster
\cite{Zhao2016}, and to robotic systems dynamically cooperating with
human operators in manufacturing and medical care \cite{Wu2019,Ton2018,Downey.Cheng.ea2017,Pi2020}.
While there is a growing demand for these applications, autonomous
agents so far have not been fully used. One major challenge is that
desired missions are often composed of multiple tasks subject to complex
specifications (e.g., complex traffic rules or sophisticated human-robot
interactions), which classical motion planning approaches, such as
the point-to-point navigation \cite{Rimon1992}, the traveling salesman
problem\cite{Halim2019}, and the orienteering problem \cite{vansteenwegen2011},
are no longer capable of. Another major challenge is that the operating
environment is often complex, e.g., dynamic and not fully known \textit{a
	priori}. The user-specified missions can be partially infeasible if
the real environment is found during the runtime to be prohibitive
to the agent. Therefore, a particular motivation for this work is
to consider online optimal motion planning of an autonomous agent
that can handle complex missions and environments.

\subsection{Related Work}

Linear temporal logic (LTL) is a formal language capable of expressing
rich task specifications and providing intuitive translation from
human language to syntactically correct formulas. Due to its rich
expressivity in describing complex missions, motion planning with
LTL specifications has generated substantial interest (cf. \cite{Belta2007,Smith2011,Guo2015,Kantaros2017,Lindemann2019,Cai2020,Cai2020optimal,Cai2020reinforcement}
to name a few). For instance, in \cite{Lacerda2019}, co-safe linear
temporal logic was used for a mobile robot to perform service tasks
in an office environment. In \cite{Cizelj2014}, standard LTL tasks
were employed for a noisy differential-drive vehicle to maximize the
probability of completing user-specified tasks in an uncertain environment.
In \cite{Schillinger2018}, task allocation and planning for temporal
logic goals were developed for a heterogeneous multi-robot system.
Other representative results on motion planning with LTL specifications
include information-guided persistent monitoring \cite{Jones2015},
hybrid control of multi-agent systems with formation constraints \cite{Guo2017}, cooperative control of mobile robots with intermittent connectivity
\cite{Kantaros2017}, and control synthesis over stochastic systems
considering both feasible and infeasible tasks \cite{Cai2020,Cai2020optimal,Cai2020reinforcement}.

Despite considerable progress in the literature, new challenges arise
when the operating environment is dynamic and uncertain. The environment
may have time-varying events of interest and dynamic obstacles that
are not fully known to the agent \textit{a priori}, which requires
the agent to dynamically adapt its motion plan to the changing environment.
To address dynamic environments, model predictive control, also referred
as receding horizon control (RHC), has been integrated with LTL specifications
and successfully applied in various applications. For instance, the
motion planning of a vehicle in an urban-like environment was considered
in \cite{Wongpiromsarn2012}, where a provably correct control strategy
that combines LTL specifications and RHC was developed. In \cite{Ding2014},
LTL-based receding horizon motion planning was developed for a finite-state
deterministic system to maximize reward collection while satisfying
desired task specifications. Recently, LTL based RHC was extended
to mobile robot networks for cooperative environmental monitoring
\cite{Lu2018}. Other results based on RHC with temporal logical specifications
include \cite{Ulusoy2014,Raman2014,Tumova2014,Farahani2017}. The
work \cite{Vasile2020} proposed a sampling-based planning algorithm
that reactively and efficiently achieved global temporal logic goals
and satisfied short-term dynamic requirements. 

Approaches based on RHC have been proven as effective tools to handle
dynamic environments in the aforementioned results. However, these
results rely on a key assumption that the operating environment is
feasible. That is, there exists a feasible motion plan in the dynamic
environment that satisfies the desired LTL specifications. However,
the assumption of a feasible dynamic environment can be restrictive,
and, in practice, not all user-specified LTL task specifications can
be realized by the agent. For instance, the agent can be tasked to
visit a sequence of areas of interest, where some of them may not
be reachable (e.g., surrounded by water that the ground robot cannot
traverse) in the real environment. LTL constraints that cannot be
fully satisfied are often relaxed to allow the tasks to be fulfilled
as much as possible. In \cite{Tumova2013},
a least-violating control strategy for finite LTL was developed to
allow potentially infeasible tasks within a partially known workspace.
In \cite{Castro2013}, sampling-based algorithm for minimum violation
motion planning was developed. In \cite{Lahijanian2015} and \cite{Lahijanian2016},
partial satisfaction of Co-safe LTL specifications was considered
to deal with uncertain environment. These strategies were further
extended in \cite{Lacerda2019} for motion planning of service robots.
In \cite{Rahmani2019optimal}, LDL$_{f}$ is applied with a graph search algorithm to treat the soft constraints.
However, only finite horizon motion planning was considered in the works of
\cite{Lacerda2019} and \cite{Tumova2013,Castro2013,Lahijanian2015,Lahijanian2016,Rahmani2019optimal}.
When considering infinite horizon motion planning, the minimal revision
problem was considered in \cite{Kim2012} and \cite{Kim2012a} with
the goal of making the revised motion planning close to the original
LTL. In \cite{Tuumova2013minimum}, minimum violation of graph-based
algorithm is presented, which is further extended in \cite{Rahmani2020}
by considering shortest plans. In \cite{Guo2015},
cooperative control synthesis of multi-agent systems was considered
in a partially known environment to maximize the satisfaction of the
specified LTL constraints. However, \cite{Kim2012,Kim2012a,Tuumova2013minimum}
focus on the single objective of minimal revision to the original
LTL. \cite{Guo2015} and \cite{Rahmani2020} optimize the static
cost (shortest path) via graph-based method, without considering motion
planning with respect to time-varying optimization objectives (e.g.,
reward collection). It is not yet understood how user-specific missions
can be successfully managed to solve optimization problems with time-varying
parameters under a dynamic environment, where desired tasks can be
partially infeasible.

\subsection{Contributions}

This work considers online motion planning of an autonomous agent
subject to LTL mission constraints. The operating environment is assumed
to be dynamic and only partially known to the agent. The environment
also has time-varying areas of interest to be visited by the agent.
The areas of interest are associated with time-varying rewards and
time-varying state labels, where the rewards indicate the relative
importance and state labels indicate time-varying workspace properties.
Since previously user-specified tasks may not be fully realized (i.e.,
partially infeasible) by the agent in the environment, this work considers
hard and soft LTL constraints, where hard constraints enforce safety
requirement (e.g. avoid obstacles) while soft constraints represents
tasks that can be relaxed to not strictly follow user-specifications
if the environment does not permit. The motion planning of the agent
is to generate policies, in decreasing order of priority, to 1) formally
guarantee the satisfaction of safety constraints; 2) mostly satisfy
soft constraints (i.e., minimize the violation cost if desired tasks
are partially infeasible); and 3) collect time-varying rewards as
much as possible (i.e., visiting areas of more interests).

To achieve these objectives, the motion of the agent is modeled by
a finite deterministic transition system (DTS), with a limited sensing
capability of detecting obstacles and observing rewards within a local
area. A relaxed product automaton is constructed based on the DTS
and the nondeterministic B\"uchi automaton (NBA) generated from the
desired LTL specifications, which allows the agent to not strictly
follow the desired LTL constraints. A utility function composed of
the violation cost and the accumulated rewards is developed, where
the violation cost is designed to quantify the differences between
the revised and the desired motion plan. The accumulated rewards are
designed to bias the motion plan towards those areas of more interests.
Since the workspace is only partially known, real-time sensed information
is used to update the agent's knowledge  about the environment. Under
the assumption of time-varying rewards that can only be locally observed,
RHC is synthesized with an LTL formula to maximize the accumulated
utilities over a finite horizon, while ensuring that safety constraints
are fully satisfied and soft constraints are mostly satisfied. 

Differing from most existing works that mainly focus
on motion planning with feasible LTL constraints, this work considers
control synthesis of an agent operating in a complex environment with
dynamic properties and time-varying areas of interest that can only
be observed locally, wherein use-specified tasks might not be fully
feasible. Integrated with the RHC framework, a relaxed product automaton
is developed to handle partially infeasible tasks by quantifying the
violation of soft constraints. RHC is synthesized with an LTL formula
to maximize the accumulated utilities over a finite horizon, while
formally ensuring the objectives in decreasing orders: 1) hard constraints
are fully satisfied; 2) soft constraints are mostly satisfied; 3)
accumulate rewards that change dynamically are locally optimized at
each time-step over finite horizon. This work is closely related to
\cite{Ding2014}. However, we extend the approach in \cite{Ding2014}
by considering partially infeasible tasks where the energy function
is redesigned to take into account the violation cost of the revised
path to the desired path. In addition, rigorous analysis is provided,
showing the correctness of the produced infinite trajectory and the
recursive feasibility of RHC-based motion planning. It's also shown
the computational complexity in automaton update is reduced. Simulation
and experiment results are provided to demonstrate its effectiveness.

\section{Preliminaries\label{sec:Pre}}

An LTL formula is built on a set of atomic propositions $\Pi$, which
are properties of system states that can be either true or false,
standard Boolean operators such as $\land$ (conjunction), $\lor$
(disjunction), $\lnot$ (negation), and temporal operators such as
$\diamondsuit$ (eventually), $\varbigcirc$ (next), $\boxempty$
(always), and $\cup$ (until). A word satisfies $\phi$ if $\phi$
is true at the first position of the word; $\boxempty\phi$ means
$\phi$ is true for all future moments; $\diamondsuit\phi$ means
$\phi$ is true at some future moments; $\varbigcirc\phi$ means $\phi$
is true at the next moment; and $\phi_{1}\mathcal{U}\phi_{2}$ means
$\phi_{1}$ is true until $\phi_{2}$ becomes true. The semantics
of an LTL formula are defined over words, which are an infinite sequence
$o=o_{0}o_{1}\ldots$ with $o_{i}\in2^{\Pi}$ for all $i\geq0$, where
$2^{\Pi}$ represents the power set of $\Pi$. Denote by $o\models\phi$
if the word $o$ satisfies the LTL formula $\phi$. More expressivity
can be achieved by combining temporal and Boolean operators. Detailed
descriptions of the syntax and semantics of LTL can be found in \cite{Clarke1999}.

An LTL formula can be translated to a nondeterministic B\"uchi automaton
(NBA).
\begin{defn}
	\label{def:NBA} An NBA is a tuple $\mathcal{B}=\left(S,S_{0},\varDelta,\Sigma,\mathcal{F}\right)$,
	where $S$ is a finite set of states; $S_{0}\subseteq S$ is the set
	of initial states; $\Sigma\subseteq2^{\Pi}$ is the input alphabet;
	$\varDelta\colon S\times\Sigma\shortrightarrow2^{S}$ is the transition
	function; and $\mathcal{F}\subseteq S$ is the set of accepting states.
\end{defn}
Let $s\overset{\sigma}{\shortrightarrow}s'$ denote the transition
from $s\in S$ to $s'\in S$ under the input $\sigma\in\Sigma$ if
$s'\in\varDelta\left(s,\sigma\right)$. Given a sequence of input
$\boldsymbol{\sigma}=\sigma_{0}\sigma_{1}\sigma_{2}\ldots$ over $\Sigma,$
a run of $\mathcal{B}$ generated by $\boldsymbol{\sigma}$ is an
infinite sequence $\boldsymbol{s}=s_{0}s_{1}s_{2}\cdots$ where $s_{0}\in S_{0}$,
and $s_{i+1}\in\varDelta\left(s_{i},\sigma_{i}\right)$ for each $i>0$.
If the input $\boldsymbol{\sigma}$ can generate at least one run
$\boldsymbol{s}$ that intersects the accepting states $\mathcal{F}$
infinitely many times, $\mathcal{B}$ is said to accept $\boldsymbol{\sigma}$.
For any LTL formula $\phi$ over $\Pi$, one can construct an NBA
with input alphabet $\Sigma=2^{\Pi}$ accepting all and only words
that satisfy $\phi$ \cite{Clarke1999}. Let $\mathcal{B}_{\phi}$
denote the NBA generated from the LTL formula $\phi$. To convert
an LTL formula to an NBA, readers are referred to \cite{Gastin2001}
for algorithms and implementations.

A dynamical system with finite states evolving deterministically under
control inputs can be modeled by a weighted finite deterministic transition
system (DTS) \cite{Baier2008}.
\begin{defn}
	\label{def:DTS} A weighted finite DTS is a tuple $\mathcal{T}=\left(Q,q_{0},\delta,\Pi,L,\omega\right)$,
	where $Q$ is a finite set of states; $q_{0}\in Q$ is the initial
	state; $\delta\subseteq Q\times Q$ is the state transitions; $\Pi$
	is the finite set of atomic propositions; $L\colon Q\shortrightarrow2^{\Pi}$
	is the labeling function; and $\omega\colon\delta\shortrightarrow\mathbb{R}^{+}$
	is the weight function.
\end{defn}
Let $q\rightarrow q'$ denote the state transition $(q,q')\in\delta$
in $\mathcal{T}$, where $q,q'\in Q$. Each transition in $\delta$
is associated with a weight determined by $\omega$. A path of $\mathcal{T}$
is an infinite sequence $\boldsymbol{q}=q_{0}q_{1}\ldots$ where $q_{i}\in Q$
and $(q_{i},q_{i+1})\in\delta$ for $i\geq0$. A path $\boldsymbol{q}$
over $\mathcal{T}$ generates an output sequence $\boldsymbol{\sigma}=\sigma_{0}\sigma_{1}\ldots$
where $\sigma_{i}=L\left(q_{i}\right)$ for $i\geq0$. The transition
$(q,q')\in\delta$ is deterministic, which implies a one-to-one map
between $\boldsymbol{q}=q_{0}q_{1}\ldots$ and the transitions $(q_{0},q_{1}),(q_{1},q_{2}),\ldots,$
thus resulting in a DTS. 

\textcolor{black}{Let $R_{k}\left(q\right)$ denote the varying reward
	associated with a state $q$ at time-step $k$. Given a trajectory
	at this time $\boldsymbol{q}_{k}=q_{0}q_{1}\ldots q_{n}$, the accumulated
	reward along the trajectory $\bar{\boldsymbol{s}}_{k}^{\mathcal{P}}$
	is $\mathbf{R}_{k}\left(\boldsymbol{q}\right)=\stackrel[i=1]{N}{\sum}R_{k}\left(q_{i}\right).$
	The time-varying reward function represents the event of interest
	in the environment}\footnote{\textcolor{black}{Local sensing rewards are considered in this work.
		Other types of rewards, such as in trajectory optimization \cite{Wolff2014},
		information gathering\cite{Leahy2015}, and local tasks \cite{Ulusoy2014},
		are also applicable.}}\textcolor{black}{. The optimization of rewards at each time-step
	is one of objectives in this paper.}

\textcolor{black}{Model checking a DTS against a LTL formula is based
	on the construction of the product automaton between the DTS and the
	corresponding NBA. Given the defined NBA and DTS, a weighted product
	automaton can be constructed as follows.}
\begin{defn}[Weighted Product Automaton]
	\textcolor{black}{\label{def:WPA} Given a weighted DTS $\mathcal{T}=\left\{ Q,q_{0},\delta,\Pi,L,\omega\right\} $
		and an NBA $\mathcal{B}=\left(S,S_{0},\varDelta,\Sigma,\mathcal{F}\right)$,
		the product automaton $\tilde{\mathcal{P}}=\mathcal{T}\times\mathcal{B}$
		is defined as a tuple $\mathcal{\tilde{\mathcal{P}}}=\left\{ P_{\mathcal{\tilde{\mathcal{P}}}},P_{\mathcal{\tilde{\mathcal{P}}}0},L_{\mathcal{\tilde{\mathcal{P}}}},\varDelta_{\mathcal{\tilde{\mathcal{P}}}},\mathcal{F}_{\mathcal{\tilde{\mathcal{P}}}},\omega_{\mathcal{\tilde{\mathcal{P}}}}\right\} $,}
	\begin{itemize}
		\item \textcolor{black}{$P_{\mathcal{\tilde{\mathcal{P}}}}=Q\times S$ is
			the set of states, e.g., $p_{\mathcal{\tilde{\mathcal{P}}}}=\left(q,s\right)$
			and $p_{\mathcal{\tilde{\mathcal{P}}}}^{\prime}=\left(q',s'\right)$
			where $p_{\mathcal{\tilde{\mathcal{P}}}},p_{\mathcal{\tilde{\mathcal{P}}}}^{\prime}\in P_{\mathcal{\tilde{\mathcal{P}}}};$}
		\item \textcolor{black}{$P_{\mathcal{\tilde{\mathcal{P}}}0}=\left\{ q_{0}\right\} \times S_{0}$
			is the set of initial states;}
		\item \textcolor{black}{$L_{\mathcal{\tilde{\mathcal{P}}}}:P_{\mathcal{\tilde{\mathcal{P}}}}\shortrightarrow2^{\Pi}$
			is a labeling function, i.e., $L_{\mathcal{\tilde{\mathcal{P}}}}\left(p_{\mathcal{\tilde{\mathcal{P}}}}\right)=L\left(q\right)$;}
		\item \textcolor{black}{$\varDelta_{\mathcal{\tilde{\mathcal{P}}}}\subseteq P_{\mathcal{\tilde{\mathcal{P}}}}\times P_{\mathcal{\tilde{\mathcal{P}}}}$
			is the set of transitions, i.e., $\left(\left(q,s\right),\left(q',s'\right)\right)\in\varDelta_{\mathcal{\tilde{\mathcal{P}}}}$
			if and only if $q\shortrightarrow q'$ and $s\overset{L(q)}{\shortrightarrow}s'$;}
		\item \textcolor{black}{$\mathcal{F}_{\mathcal{\tilde{\mathcal{P}}}}=Q\times\mathcal{F}$
			is the set of accepting states;}
		\item \textcolor{black}{$\omega_{\mathcal{\tilde{\mathcal{P}}}}\colon\varDelta_{\mathcal{\tilde{\mathcal{P}}}}\shortrightarrow\mathbb{R}^{+}$
			is the weight function, i.e., $\omega_{\mathcal{\tilde{\mathcal{P}}}}\left(p_{\mathcal{\tilde{\mathcal{P}}}},p_{\mathcal{\tilde{\mathcal{P}}}}^{\prime}\right)=\omega\left(q,q'\right)$.}
	\end{itemize}
\end{defn}
\textcolor{black}{Let $\left(q,s\right)\rightarrow_{\tilde{\mathcal{P}}}\left(q',s'\right)$
	denote the transitions from $\left(q,s\right)=s_{\mathcal{\tilde{\mathcal{P}}}}$
	to $\left(q',s'\right)=p_{\mathcal{\tilde{\mathcal{P}}}}^{\prime}$
	in $\tilde{\mathcal{P}}$ if $\left(p_{\mathcal{\tilde{\mathcal{P}}}},p_{\mathcal{\tilde{\mathcal{P}}}}^{\prime}\right)\in\varDelta_{\mathcal{\tilde{\mathcal{P}}}}$.
	A trajectory $\boldsymbol{p}_{\tilde{\mathcal{P}}}=\left(q_{0},s_{0}\right)\left(q_{1},s_{1}\right)\ldots$
	of $\tilde{\mathcal{P}}$ is an infinite sequence where $\left(q_{0},s_{0}\right)\in S_{\mathcal{\tilde{\mathcal{P}}}0}$
	and $\left(q_{i},s_{i}\right)\rightarrow_{\tilde{\mathcal{P}}}\left(q_{i+1},s_{i+1}\right)$
	for all $i\geq0$. The trajectory $\boldsymbol{p}_{\tilde{\mathcal{P}}}$
	is called accepting if and only if $\boldsymbol{s}_{\tilde{\mathcal{P}}}$
	intersects $\mathcal{F}_{\mathcal{\tilde{\mathcal{P}}}}$ infinitely
	many times. Let $\gamma_{\mathcal{T}}\left(\boldsymbol{p}_{\tilde{\mathcal{P}}}\right)=q_{0}q_{1}\ldots$
	denote the projection of $\boldsymbol{p}_{\tilde{\mathcal{P}}}$ on
	the transition system $\mathcal{T}$. Note that a trajectory $\boldsymbol{s}_{\tilde{\mathcal{P}}}$
	can be uniquely projected onto $\mathcal{T}$ by $\gamma_{\mathcal{T}}$.
	By the construction of $\tilde{\mathcal{P}}$ from $\mathcal{T}$
	and $\mathcal{B}$, $\boldsymbol{s}_{\tilde{\mathcal{P}}}$ is an
	accepting trajectory on $\mathcal{\tilde{\mathcal{P}}}$ if and only
	if $\gamma_{\mathcal{T}}\left(\boldsymbol{p}_{\tilde{\mathcal{P}}}\right)$
	satisfies the LTL formula corresponding to $\mathcal{B}$. }

\section{Example and Problem Formulation}\label{subsec:problem-formulation}

\subsection{Example Demonstration}

\begin{figure}
\centering{}\includegraphics[scale=0.27]{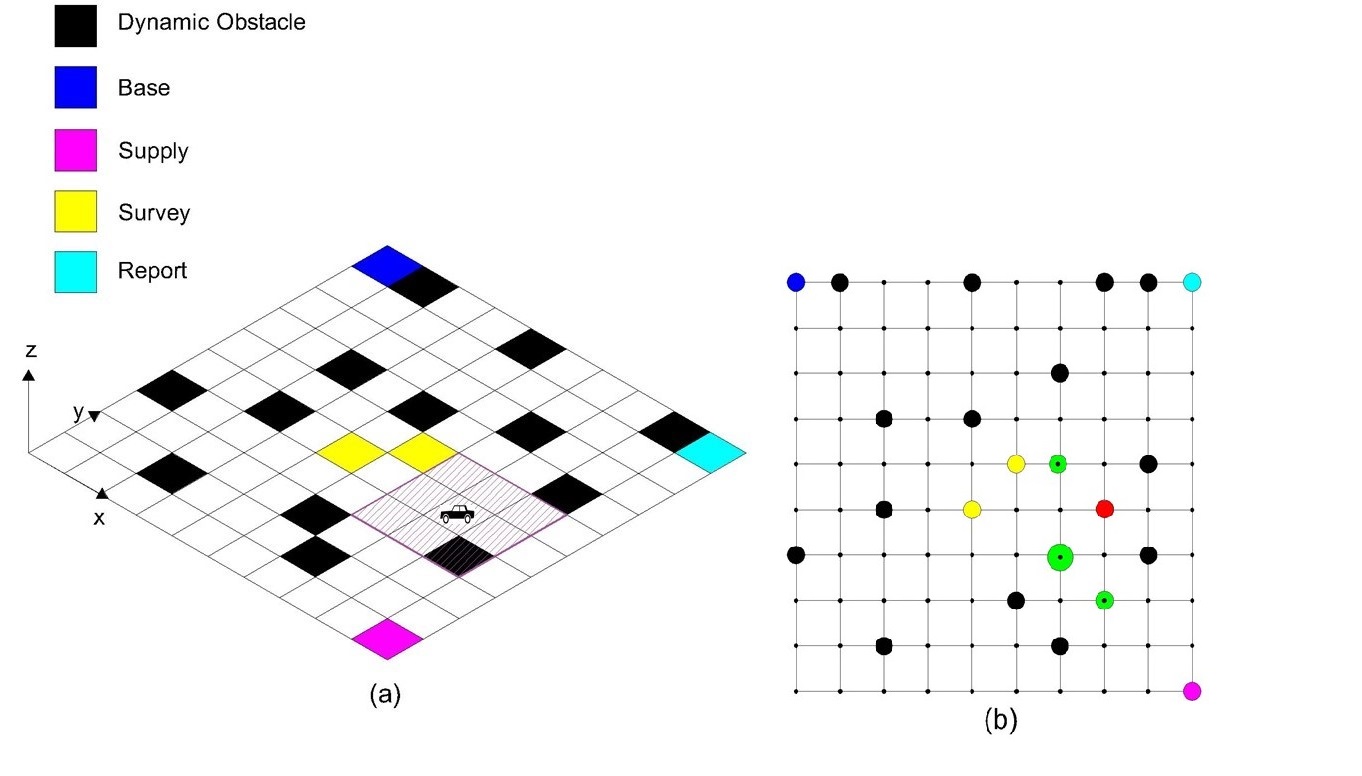}\caption{\label{fig:scenario}  (a) Example of a partitioned operating environment,
	where the shaded area around the vehicle indicates its local sensing.
	(b) The corresponding abstracted grid-like graph of (a), where the
	size of green dots is proportional to their reward values and the
	red dot represents the vehicle. }
\end{figure}
As a running example, consider a robot operating
in an environment abstracted to a labeled grid-like graph $\mathcal{G}=\left(\mathcal{V},\mathcal{E},\Pi\right)$,
where the node set $\mathcal{V}$ represents the partitioned areas,
the edge set $\mathcal{E}$ indicates possible transitions, and the
atomic propositions $\Pi=\{\mathtt{\mathtt{Base}},\thinspace\mathtt{Supply},\thinspace\mathtt{Report},\thinspace\mathtt{Obstacle},\thinspace\mathtt{Survey}\}$
indicate the labeled properties of the areas\footnote{Abstracted environments have been widely used in
	the literature, and many existing partition methods, such as triangulation
	and rectangular grids, can be applied to partition the workspace \cite{Kloetzer2008}.}, as shown in Fig. \ref{fig:scenario}. The robot
motion in the environment is then represented by the finite DTS $\mathcal{T}$
in Def. \ref{def:DTS} evolving over $\mathcal{G}$, where $Q$ represents
the node set $\mathcal{V}$, and the possible transitions $\delta$
are captured by the edge set $\mathcal{E}.$ As an example application,
a specific surveillance mission $\phi$ is considered in this work,
where the robot is required to visit a set of stations repetitively,
while maximizing the collected rewards and avoiding obstacles on the
way to the destinations. Due to the consideration of partially infeasible
environment, the user-specified LTL task $\phi$ consists of hard
constraints $\phi_{h}$ and soft constraints $\phi_{s}$, i.e., $\phi=\phi_{h}\land\phi_{s}$,
where $\phi_{h}$ models the constraint of collision avoidance that
have to be fully satisfied while $\phi_{s}$ models soft constraints
that can be relaxed if it's infeasible in current environment. In
this case, The desired task of the robot within the environment $\mathcal{G}$
is described by an LTL formula $\phi$ over the atomic propositions
$\Pi$. A variety of tasks can be represented in LTL formulas, such
as the sequential visit of $\mathtt{Survey}$ and $\mathtt{Report}$
(i.e., $\diamondsuit\left(\mathtt{Survey}\land\diamondsuit\mathtt{Report}\right)$),
the persistent surveillance of visiting $\mathtt{\mathtt{Base}}$
infinitely many times (i.e., $\boxempty\,\diamondsuit\mathtt{\mathtt{Base}}$),
and avoiding collision while achieving a task $\phi$ (i.e., $\boxempty\left(\lnot\mathtt{Obstacle}\land\phi\right)$).
More expressivity of tasks can be achieved based on the combination
of temporal and Boolean operators over $\Pi$. The environment is
assumed to be only partially known to the robot, i.e., the robot may
know the static destinations to visit but not the obstacles it may
encounter during mission operation. The environment is dynamic in
the sense of containing real-time dynamic obstacles and time-varying
rewards associated with each state. The time-varying reward $R_{k}\left(q\right)\in\mathbb{R}^{+}$
is given at each step. It is further assumed that the robot can only
detect obstacles, observe rewards, and sense node labels within a
local area around itself. As an example application, the robot is
required to complete the given LTL task, while maximizing the collected
rewards. More detailed task descriptions can be found in Section \ref{subsec:Simu}.

The description above is just an representation
and this work can be extended in several directions. The LTL tasks
can be defined for other sets of atomic propositions and requirements
and the rewards-optimization problem can be easily applied to other
meaningful time-varying objectives. 

\subsection{Problem formulation}

Given the time-varying reward function $R_{k}\left(q_{i}\right)$,
	$\forall i=1,\ldots,N$ associated with each states in DTS that is
	unknown a priori, and is only observed and optimised locally online
	at time $k$, the motion planning problem in this work is presented
	as follows. 
\begin{problem}
	\label{Prob1}Given a deterministic transition system
		$\mathcal{T}$, and a user-specified LTL formula $\phi=\phi_{h}\land\phi_{s}$,
		the control objective is to design an online planning strategy, in
		decreasing order of priority, that 1) $\phi_{h}$ is fully satisfied;
		2) $\phi_{s}$ is fulfilled as much as possible if the task is not
		feasible; and 3) rewards collection at each time-step is maximized
		over a finite horizon during mission operation.
\end{problem}
In Problem \ref{Prob1}, by saying to fulfill $\phi_{s}$
as much as possible, we mean to minimize the violation of $\phi_{s}$,
which will be formally defined in Section \ref{subsec:relax}.

\section{Relaxed Automaton and Problem Formulation\label{sec:RELAXED-AUTOMATON and PROBLEM-FORMULATION}}

Sec. \ref{subsec:relax} discusses how $\phi_{s}$ can be relaxed
to allow motion revision and how the violation of $\phi_{s}$ can
be quantified. Sec. \ref{subsec:Energy} describes the construction
of an energy function that enforces the satisfaction of accepting
conditions. Sec. \ref{subsec:Automaton} presents how local sensing
can be used to update the robot's knowledge about the environment
to facilitate motion revision.

\subsection{Relaxed LTL Specifications\label{subsec:relax}}

Let $\mathcal{B}_{h}=\left(S_{h},S_{h0},\varDelta_{h},\Sigma_{h},\mathcal{F}_{h}\right)$
and $\mathcal{B}_{s}=\left(S_{s},S_{s0},\varDelta_{s},\Sigma_{s},\mathcal{F}_{s}\right)$
denote the NBA corresponding to $\phi_{h}$ and $\phi_{s}$, respectively.
The relaxed product automaton for $\phi=\phi_{h}\land\phi_{s}$ is
constructed as follows. 
\begin{defn}[Relaxed Product Automaton]
	\label{def:RPA} Given a weighted DTS $\mathcal{T}=\left\{ Q,q_{0},\delta,\Pi,L,\omega\right\} $
	and the NBA $\mathcal{B}_{h}$ and $\mathcal{B}_{s}$, the relaxed
	product automaton $\mathcal{P}=\mathcal{T}\times\mathcal{B}_{h}\times\mathcal{B}_{s}$
	is defined as a tuple $\mathcal{P}=\mathcal{T}\times\mathcal{B}_{h}\times\mathcal{B}_{s}$
	is defined as a tuple $\mathcal{P}=\left\{ S_{\mathcal{P}},S_{\mathcal{P}0},L_{\mathcal{P}},\varDelta_{\mathcal{P}},\mathcal{F}_{\mathcal{P}},\omega_{\mathcal{P}},h_{\mathcal{P}},\mathrm{v_{\mathcal{P}}}\right\} $,
	where
	\begin{itemize}
		\item $S_{\mathcal{P}}=Q\times S_{h}\times S_{s}$ is the set of states,
		e.g., $s_{\mathcal{P}}=\left(q,s_{h},s_{s}\right)$ and $s_{\mathcal{P}}^{\prime}=\left(q',s_{h}',s_{s}'\right)$
		where $s_{\mathcal{P}},s_{\mathcal{P}}^{\prime}\in S_{\mathcal{P}}$;
		\item $S_{\mathcal{P}0}=\left\{ q_{0}\right\} \times S_{h0}\times S_{s0}$
		is the set of initial states;
		\item $L_{\mathcal{P}}:S_{\mathcal{P}}\rightarrow2^{\Pi}$ is a labeling
		function, i.e., $L_{\mathcal{P}}\left(s_{\mathcal{P}}\right)=L\left(q\right)$;
		\item $\varDelta_{\mathcal{P}}\subseteq S_{\mathcal{P}}\times S_{\mathcal{P}}$
		is the set of transitions, i.e., defined by $\left(\left(q,s_{h},s_{s}\right),\left(q',s_{h}',s_{s}'\right)\right)\in\varDelta_{\mathcal{P}}$
		if and \textcolor{black}{only if $(q,q')\in\delta$, $\exists l_{h}\in2^{\Pi_{h}}$
			and $\exists l_{s}\in2^{\Pi_{s}}$ such that $s_{h}'\in\varDelta\left(s_{h},l_{h}\right)$
			and $s_{s}'\in\varDelta\left(s_{s},l_{s}\right)$;}
		\item \textcolor{black}{$h_{\mathcal{P}}\colon\varDelta_{\mathcal{P}}\rightarrow\left\{ 0,\infty\right\} $
			;}
		\item $\omega_{\mathcal{P}}\colon\varDelta_{\mathcal{P}}\rightarrow\mathbb{R}^{+}$
		is the weight function;
		\item $\mathrm{v_{\mathcal{P}}}\colon\varDelta_{\mathcal{P}}\rightarrow\mathbb{R}^{+}$
		is the violation function;
		\item $\mathcal{F}_{\mathcal{P}}=Q\times\mathcal{F}_{h}\times\mathcal{F}_{s}$
		is the set of accepting states.
	\end{itemize}
	\textcolor{black}{The major difference between $\tilde{\mathcal{P}}$
		and $\mathcal{P}$ is that for two any state $s_{\mathcal{P}}=\left(q,s_{h},s_{s}\right)$
		and $s_{\mathcal{P}}^{\prime}=\left(q',s_{h}',s_{s}'\right)$, the
		constraints $s_{h}'\in\varDelta\left(s_{h},L\left(q\right)\right)$
		and $s_{s}'\in\varDelta\left(s_{s},L\left(q\right)\right)$ in $\mathcal{\tilde{\mathcal{P}}}$
		are relaxed in $\mathcal{P}$ as defined above. Consequently, $\mathcal{P}$
		is more connected than $\tilde{\mathcal{P}}$ in terms of possible
		transitions, which will reduce the computational complexity during
		automaton update (see Section \ref{subsec:Automaton}). Any transition
		$\left(\left(q,s_{h},s_{s}\right),\left(q,s_{h}',s_{s}'\right)\right)\in\varDelta_{\mathcal{P}}$
		that violates the hard constraint will have a infinite violation $h_{\mathcal{P}}\left(\left(\left(q,s_{h},s_{s}\right),\left(q,s_{h}',s_{s}'\right)\right)\right)$.
		To identify trajectories that violate the original $\phi_{s}$ the
		least when the environment is infeasible, $\mathrm{v_{\mathcal{P}}}$
		is designed to quantify the violation cost. Suppose that $\Pi=\left\{ \alpha_{1},\alpha_{2}\ldots\alpha_{M}\right\} $
		and consider an evaluation function $\Eval\colon2^{\Pi}\shortrightarrow\left\{ 0,1\right\} ^{M}$,
		where $\Eval\left(l\right)=\left(v_{i}\right){}^{M}$ with $v_{i}=1$
		if $\alpha_{i}\in l$ and $v_{i}=0$ if $\alpha_{i}\notin l$, where
		$i=1,2,\ldots,M$ and $l\in2^{\Pi}$. To quantify the difference between
		two elements in $2^{\Pi}$, consider $\rho\left(l,l'\right)=\left\Vert v-v^{\prime}\right\Vert _{1}=\sum_{i=1}^{M}\left|v_{i}-v_{i}^{\prime}\right|,$
		where $v=\Eval\left(l\right)$, $v^{\prime}=\Eval\left(l^{\prime}\right)$,
		$l,l^{\prime}\in2^{\Pi}$, and $\left\Vert \cdot\right\Vert _{1}$
		is the $l_{1}$ norm. The distance from $l\in2^{\Pi}$ to a set $\mathcal{X}\subseteq2^{\Pi}$
		is then defined as $\Dist\left(l,\ensuremath{\mathcal{X}}\right)=\underset{l'\in\ensuremath{\mathcal{X}}}{\min}\rho\left(l,l'\right)$
		if $\text{ }l\notin\ensuremath{\mathcal{X}}$, and $\Dist\left(l,\ensuremath{\mathcal{X}}\right)=0$
		otherwise. Now the violation cost of the transition from $s_{\mathcal{P}}=\left(q,s_{h},s_{s}\right)$
		to $s_{\mathcal{P}}^{\prime}=\left(q',s_{h}',s_{s}'\right)$ can be
		defined as ${\color{blue}{\color{black}\mathrm{v_{\mathcal{P}}}\left(s_{\mathcal{P}},s_{\mathcal{P}}^{\prime}\right)=\Dist\left(L\left(q\right),\ensuremath{\mathcal{X}}\left(s_{s},s_{s}'\right)\right)}}$,
		where $\ensuremath{\mathcal{X}}\left(s_{s},s_{s}^{\prime}\right)=\left\{ l\in2^{\Pi}\left|s_{s}'\in\varDelta\left(s_{s},l\right)\right.\right\} $
		is the set of input alphabets that enables the transition from $s_{s}$
		to $s_{s}^{\prime}$. Hence, the violation cost $\mathrm{v_{\mathcal{P}}}\left(s_{\mathcal{P}},s_{\mathcal{P}}^{\prime}\right)$
		quantifies how much the transition from $s_{\mathcal{P}}$ to $s_{\mathcal{P}}^{\prime}$
		in $\mathcal{P}$ violates the constraints imposed by $\phi_{s}$. }
\end{defn}
\textcolor{black}{Based on the defined $\mathrm{v_{\mathcal{P}}}\left(s_{\mathcal{P}},s_{\mathcal{P}}^{\prime}\right)$,
	we design the weight function $\omega_{\mathcal{P}}\left(s_{\mathcal{P}},s_{\mathcal{P}}^{\prime}\right)=h_{\mathcal{P}}\left(s_{\mathcal{P}},s_{\mathcal{P}}^{\prime}\right)+\omega\left(q,q'\right)+\beta\mathrm{\cdot v_{\mathcal{P}}}\left(s_{\mathcal{P}},s_{\mathcal{P}}^{\prime}\right)$,
	where $\beta\in\mathbb{R}^{+}$ indicates the relative penalty. A
	larger $\beta$ tends to bias the selection of trajectories with less
	violation cost. The weight function $\omega\left(q,q'\right)$ is
	defined on the Euclidean distance between $q$ and $q'$ on $\mathcal{T}$,
	which measures the implementation cost of the transition from $q$
	to $q'$. Since each transition $\left(s_{k}^{\mathcal{P}},s_{k+1}^{\mathcal{P}}\right)\in\varDelta_{\mathcal{\mathcal{P}}}$
	is associated with a weight in Def. \ref{def:RPA}, the total weight
	of a trajectory $\boldsymbol{s}_{\mathcal{P}}$ is 
	\begin{equation}
	{\begin{alignedat}{1}\mathcal{W}\left(\boldsymbol{s}_{\mathcal{P}}\right)= & \sum_{k=1}^{n-1}\left(h_{\mathcal{P}}\left(s_{k}^{\mathcal{P}},s_{k+1}^{\mathcal{P}}\right)+\omega\left(q_{k},q_{k+1}\right)\right.\\
		& \left.+\beta\cdot\ensuremath{\mathrm{v_{\mathcal{P}}}}\left(s_{k}^{\mathcal{P}},s_{k+1}^{\mathcal{P}}\right)\right).
		\end{alignedat}
	}\label{eq:total-cost}
	\end{equation}
}
\begin{thm}
	\label{Thm:hard_components} Given an accepting run $\boldsymbol{s}_{\mathcal{P}}=\left(q_{0},s_{h0},s_{s0}\right)\left(q_{1},s_{h1},s_{s1}\right)\ldots$
	of $\mathcal{P}$ for $\phi=\phi_{h}\land\phi_{s}$, the hard constraints
	$\phi_{h}$ will always be satisfied if $\mathcal{W}\left(\boldsymbol{s}_{\mathcal{P}}\right)\neq\infty$. 
\end{thm}
\begin{IEEEproof}
	Let $\mathcal{F}_{h}$ denote the set of accepting states of $\mathcal{B}_{h}$
	corresponding to $\mathcal{F}_{\mathcal{P}}$ (i.e., the projection
	of $\mathcal{F}_{\mathcal{P}}$ of $\mathcal{P}$ onto $\mathcal{F}_{h}$
	of $\mathcal{B}_{h}$), and let $\boldsymbol{s_{h}}=s_{h0}s_{h1}\ldots$
	denote the projection of $\boldsymbol{s}_{\mathcal{P}}$ over $\mathcal{P}$
	onto $\mathcal{B}_{h}$. By the definition of an accepting run, $\boldsymbol{s}_{\mathcal{P}}$
	intersects at least one state of $\mathcal{F}_{\mathcal{P}}$ infinitely
	often, which implies $s_{h}$ visits $\mathcal{F}_{h}$ infinitely
	often. In addition, by the definition of $\varDelta_{\mathcal{P}}\subseteq S_{\mathcal{P}}\times S_{\mathcal{P}}$,
	all transitions along $\boldsymbol{s_{h}}$ follow the rul\textcolor{black}{e
		$s_{hi}\stackrel{L\left(q\right)}{\rightarrow}s_{hj}$ if $h_{\mathcal{P}}\left(\boldsymbol{s}_{\mathcal{P}i},\boldsymbol{s}_{\mathcal{P}j}\right)=0$,
		w}hich implies the transitions are always valid in $\mathcal{B}_{h}$.
	Therefore, $s_{h}$ satisfies the accepting conditions of $\mathcal{B}_{h},$
	which implies that $\phi_{h}$ is fully satisfied. 
\end{IEEEproof}
\textcolor{black}{Theorem \ref{Thm:hard_components} indicates that
	any accepting run of $\mathcal{P}$ can guarantee that the hard constraints
	$\phi_{h}$ are satisfied by selecting the run with finite total cost
	in (\ref{eq:total-cost}). An accepting run $\boldsymbol{s}_{\mathcal{P}}$
	is valid if and only if it satisfies $\phi_{h}$. In (\ref{eq:total-cost}),
	the term $\sum_{k=1}^{n-1}\beta\cdot\mathrm{v_{\mathcal{P}}}\left(s_{k}^{\mathcal{P}},s_{k+1}^{\mathcal{P}}\right)$
	measures the violation of $\phi_{s}$. Hence, a valid accepting run
	$\boldsymbol{s}_{\mathcal{P}}$ fulfills $\phi_{s}$ as much as possible,
	if the violation of $\phi_{s}$ can be minimized. }
\begin{example}
\begin{figure}
\centering{}\includegraphics[scale=0.35]{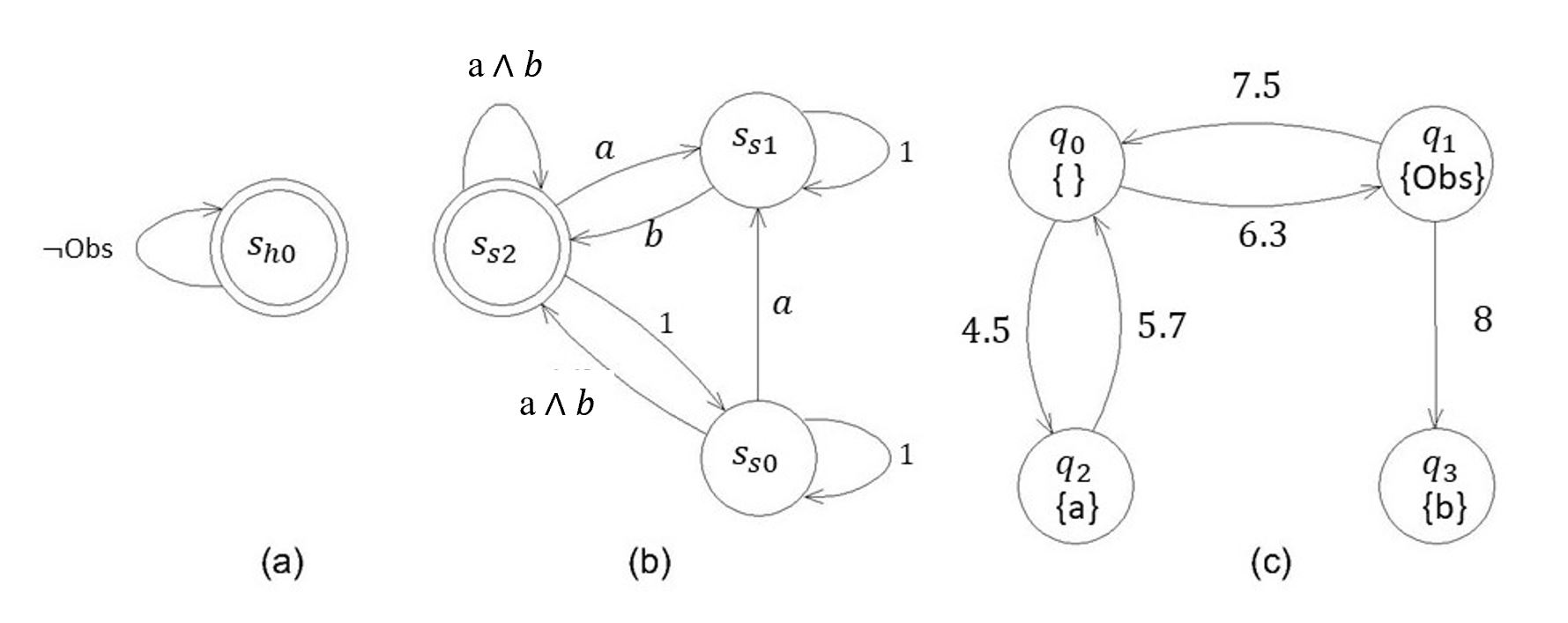}\caption{\label{fig:Buchi and Ts} (a) Example of safety constraint $\phi_{h}=\protect\lnot Obs$.
(b) Example of a soft constraint $\phi_{s}=\boxempty\diamondsuit a\lor\boxempty\diamondsuit b$.
(c) Example of DTS.}
\end{figure}
\begin{figure}
\centering{}\includegraphics[scale=0.37]{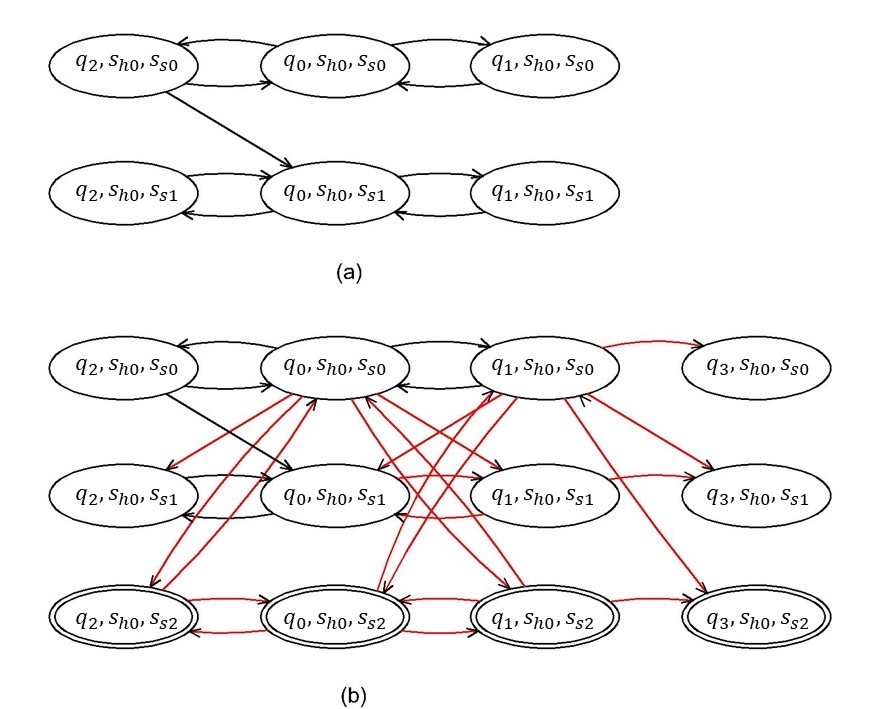}\caption{\label{fig:product automaton} (a) Product automaton based on Def.\ref{def:WPA}.
(b) Relaxed product automaton based on Def.\ref{def:RPA}.}
\end{figure}
 Consider an example of $\phi=\phi_{h}\land\phi_{s}$ and $\mathcal{T}$
in Fig. \ref{fig:Buchi and Ts}. The double circles in Fig. \ref{fig:Buchi and Ts}(a)
and (b) represent the accepting states of $\mathcal{B}_{h}$ and $\mathcal{B}_{s}$,
respectively. Fig. \ref{fig:Buchi and Ts}(a) represents safety constraints
of avoiding obstacles and Fig. \ref{fig:Buchi and Ts}(b) represents
soft constraints of visiting $a$ and $b$ infinitely often. Fig.
\ref{fig:Buchi and Ts}(c) represents a DTS with labeled states. Fig.
\ref{fig:product automaton}(a) shows the product automaton in Def.
\ref{def:WPA} between $\mathcal{T}$ and the two NAB $\phi_{h}$
and $\phi_{s}$. Fig. \ref{fig:product automaton}(b) shows the relaxed
product automaton in Def. \ref{def:RPA}, where red edges represent
transitions with non-zero violation cost. The plot in Fig. \ref{fig:product automaton}
omits the states that are not reachable from initial states and double
circle states are accepting states. In Fig. \ref{fig:product automaton}(a),
there exists no accepting path since accepting states are not reachable
due to constraints of $\phi_{h}$. In contrast, in the relaxed automaton,
there exists a path visiting accepting states infinitely often. In
this case, the projection of accepting path with minimum violation
cost in Fig. \ref{fig:product automaton}(b) onto DTS $\mathcal{T}$
is the path visiting $a$ of Fig. \ref{fig:Buchi and Ts}(c) infinitely
often while avoiding obstacles. 
\end{example}
\begin{rem}
	Since the accepting condition of $\mathcal{P}$ requires to visit
	both accepting states of $\mathcal{B}_{h}$ and $\mathcal{B}_{s}$
	infinitely, there should exist no conflicts between the satisfaction
	of $\mathcal{B}_{h}$ and $\mathcal{B}_{s}$ in this work and \cite{Cai2020Letters}.
	The more general construction without considering such conflicts can
	be found in\textcolor{black}{{} \cite{Guo2015} resulting extra dimensional
		complexity and our framework also works for the general design.}
\end{rem}

\subsection{Energy Function\label{subsec:Energy}}

Analogous to Lyapunov theory, where the convergence
of the system states to equilibrium points is indicated by a decreasing
Lyapunov function, a Lyapunov-like energy function is designed in
this section to enforce the acceptance condition of an automaton by
requiring the distance to the accepting states to decrease as the
system evolves. Given two states $r_{i}^{\mathcal{P}},s_{j}^{\mathcal{P}}\in S_{\mathcal{P}}$,
the set of all finite trajectories on $\mathcal{P}$ from $s_{i}^{\mathcal{P}}$
to $s_{j}^{\mathcal{P}}$ is defined as $\mathcal{D}\left(s_{i}^{\mathcal{P}},s_{j}^{\mathcal{P}}\right)=\left\{ \boldsymbol{s}_{\mathcal{P}}=s_{1}^{\mathcal{P}}s_{2}^{\mathcal{P}}\ldots s_{n}^{\mathcal{P}}\right\} $
such that $s_{1}^{\mathcal{P}}=s_{i}^{\mathcal{P}}$, $s_{n}^{\mathcal{P}}=s_{j}^{\mathcal{P}}$,
$\left(s_{k}^{\mathcal{P}},s_{k+1}^{\mathcal{P}}\right)\in\varDelta_{\mathcal{\mathcal{P}}}$,
$\forall k=1,2\ldots,n-1$. If $\mathcal{D}\left(s_{i}^{\mathcal{P}},s_{j}^{\mathcal{P}}\right)\neq\emptyset$,
$s_{j}^{\mathcal{P}}$ is reachable from $s_{i}^{\mathcal{P}}$ on
$\mathcal{P}$. Based on (\ref{eq:total-cost}), the distance $d\left(s_{i}^{\mathcal{P}},s_{j}^{\mathcal{P}}\right)$
is defined as the lowest total weight along a trajectory from $s_{i}^{\mathcal{P}}$
to $s_{j}^{\mathcal{P}}$, i.e.,
\begin{equation}
d\left(s_{i}^{\mathcal{P}},s_{j}^{\mathcal{P}}\right)=\left\{ \begin{array}{cc}
\underset{\boldsymbol{s}_{\mathcal{P}}\in\ensuremath{\mathcal{D}\left(s_{i}^{\mathcal{P}},s_{j}^{\mathcal{P}}\right)}}{\min}\mathcal{W}\left(\boldsymbol{s}_{\mathcal{P}}\right) & \text{ if }\mathcal{D}\left(s_{i}^{\mathcal{P}},s_{j}^{\mathcal{P}}\right)\neq\emptyset,\\
\infty & \text{Otherwise. }
\end{array}\right.\label{eq:LowestPath}
\end{equation}
$d\left(s_{i}^{\mathcal{P}},s_{j}^{\mathcal{P}}\right)$ can be efficiently
determined by the well known Dijkstra's algorithm. 

\textcolor{black}{Given $\mathcal{P}_{\left(S_{\mathcal{P}},\varDelta_{\mathcal{P}}\right)}$
, the graph induced from $\mathcal{P}_{\left(S_{\mathcal{P}},\varDelta_{\mathcal{P}}\right)}$
by neglecting the weight of each transition is denoted by $\mathcal{G}_{\left(S_{\mathcal{P}},\varDelta_{\mathcal{P}}\right)}.$ }
\begin{defn}
\textcolor{black}{\label{def:self-reachable sets}}\textcolor{blue}{{}
}\textcolor{black}{The largest self-reachable subset of the accepting
	set $\mathcal{F}_{\mathcal{P}}$ is defined as $\mathcal{\mathcal{F}}^{*}$
	such that each pair states of $\mathcal{\mathcal{F}}^{*}$ can reach
	each other in $\mathcal{P}$.}
\end{defn}
\textcolor{black}{$\mathcal{\mathcal{F}}^{*}$ in this paper can be
constructed by following similar procedures in \cite{Ding2014} by
neglecting the cost of $h_{\mathcal{P}}$. }
\begin{defn}[Energy Function]
\textcolor{black}{\label{def:energy} For $s_{\mathcal{P}}\in S_{\mathcal{P}}$,
	the energy function $J\left(s_{\mathcal{P}}\right)$ is designed as}

\textcolor{black}{
	\begin{equation}
	J\left(s_{\mathcal{P}}\right)=\left\{ \begin{array}{cc}
	\underset{s_{\mathcal{P}}^{\prime}\in\mathcal{F}^{*}}{\min}d\left(s_{\mathcal{P}},s_{\mathcal{P}}^{\prime}\right) & \text{ if }s_{\mathcal{P}}\notin\mathcal{F}^{*},\\
	0 & \text{ if }s_{\mathcal{P}}\in\mathcal{F}^{*}.
	\end{array}\right.\label{eq:energy}
	\end{equation}
}
\end{defn}
\textcolor{black}{The design of $J\left(s_{\mathcal{P}}\right)$ in
(\ref{def:energy}) is inspired by \cite{Ding2014}. Different from
\cite{Ding2014}, we adapt it to the relaxed product automaton by
taking into account the distance from the states to the largest self-reachable
subset $\mathcal{\mathcal{F}}^{*}$ of the relaxed product automaton.
Since $\omega_{\mathcal{P}}$ is positive by definition, $d\left(s_{\mathcal{P}},s_{\mathcal{P}}^{\prime}\right)>0$
for all $s_{\mathcal{P}},s_{\mathcal{P}}^{\prime}\in S_{\mathcal{P}}$,
which implies that $J\left(s_{\mathcal{P}}\right)\geq0$. Particularly,
$J\left(s_{\mathcal{P}}\right)=0$ if $s_{\mathcal{P}}\in\mathcal{F}^{*}$.
If a state in $\mathcal{F}^{*}$ is reachable from $s_{\mathcal{P}}$,
then $J\left(s_{\mathcal{P}}\right)\neq\infty$, otherwise $J\left(s_{\mathcal{P}}\right)=\infty$.
Hence, $J\left(s_{\mathcal{P}}\right)$ indicates the minimum distance
from $s_{\mathcal{P}}$ to $\mathcal{F}^{*}$. }
\begin{thm}
\label{Thm:energy}For the energy function designed in (\ref{eq:energy}),
if a trajectory $\boldsymbol{s}_{\mathcal{P}}=s_{1}^{\mathcal{P}}s_{2}^{\mathcal{P}}\ldots s_{n}^{\mathcal{P}}$
is accepting, there is no state $s_{i}^{\mathcal{P}}$, $\forall i=1,\ldots,n$,
with $J\left(s_{i}^{\mathcal{P}}\right)=\infty$, and all accepting
states in $\boldsymbol{s}_{\mathcal{P}}$ are in the set $\mathcal{F}^{*}$
with energy 0. In addition, for any state $s_{\mathcal{P}}\in S_{\mathcal{P}}$
with $s_{\mathcal{P}}\notin\mathcal{F}^{*}$ and $J\left(s_{\mathcal{P}}\right)\neq\infty$,
there exists at least one state $s_{\mathcal{P}}^{\prime}$ with $\left(s_{\mathcal{P}},s_{\mathcal{P}}^{\prime}\right)\in\varDelta_{\mathcal{P}}$
such that $J\left(s_{\mathcal{P}}^{\prime}\right)<J\left(s_{\mathcal{P}}\right)$.
\end{thm}
\begin{IEEEproof}
Consider an accepting state $s_{i}^{\mathcal{P}}\in\mathcal{F}_{\mathcal{P}}$.
Suppose $s_{i}^{\mathcal{P}}\notin\mathcal{F}^{*}$. If a trajectory
$\boldsymbol{s}_{\mathcal{P}}$ is accepting, $\boldsymbol{s}_{\mathcal{P}}$
must intersect $\mathcal{F}_{\mathcal{P}}$ infinitely many times
by Def. \ref{def:RPA}, which indicates there exists another state
$s_{j}^{\mathcal{P}}\in\mathcal{F}_{\mathcal{P}}$ such that $s_{j}^{\mathcal{P}}$
is reachable from $s_{i}^{\mathcal{P}}$. If $s_{j}^{\mathcal{P}}\in\mathcal{F}^{*}$,
the construction of $\mathcal{F}^{*}$ indicates that $s_{i}^{\mathcal{P}}$
must be in $\mathcal{F}^{*},$ which contradicts the assumption that
$s_{i}^{\mathcal{P}}\notin\mathcal{F}^{*}$. If $s_{j}^{\mathcal{P}}\notin\mathcal{F}^{*},$
there must exist a non-trivial strongly connected component (SCC)
composed of accepting states reachable from $s_{j}^{\mathcal{P}}$
\cite{Baier2008}. By definition of $\mathcal{F}^{*}$, all states
in SCC belong to $\mathcal{F}^{*}$. Since the SCC is reachable from
$s_{j}^{\mathcal{P}}$, it implies $s_{j}^{\mathcal{P}}\in\mathcal{F}^{*}$,
which contradicts that $s_{j}^{\mathcal{P}}\notin\mathcal{F}^{*}$.
Consequently, all accepting states in $\boldsymbol{s}_{\mathcal{P}}$
must be in $\mathcal{F}^{*}$ and have energy zero based on (\ref{eq:energy}).
Since $\mathcal{F}^{*}$ is reachable by any state in $\boldsymbol{s}_{\mathcal{P}}$,
$J\left(s_{i}^{\mathcal{P}}\right)\neq\infty$ for all $i=1,\ldots,n$.

If $J\left(s_{\mathcal{P}}\right)\neq\infty$ for $s_{\mathcal{P}}\in S_{\mathcal{P}}$,
(\ref{eq:energy}) indicates $\mathcal{F}^{*}$ is reachable strictly
following the hard constraint from $s_{\mathcal{P}}$. That is, based
on the distance defined in (\ref{eq:LowestPath}), there exists a
shortest trajectory $\boldsymbol{s}_{\mathcal{P}}=s_{1}^{\mathcal{P}}s_{2}^{\mathcal{P}}\ldots s_{n}^{\mathcal{P}}$,
where $s_{1}^{\mathcal{P}}=s_{\mathcal{P}}$ and $s_{n}^{\mathcal{P}}\in\mathcal{F}^{*}$.
Bellman's optimal principle can then be invoked to conclude that there
exists a state $s_{\mathcal{P}}^{\prime}$ with $\left(s_{\mathcal{P}},s_{\mathcal{P}}^{\prime}\right)\in\varDelta_{\mathcal{P}}$
such that $J\left(s_{\mathcal{P}}^{\prime}\right)<J\left(s_{\mathcal{P}}\right)$.
\end{IEEEproof}
Theorem \ref{Thm:energy} indicates that, as long as the energy function
keeps decreasing, the generated trajectory will eventually satisfy
the accepting conditions in Def. \ref{def:RPA}. As a result, the
designed energy function can be used to enforce the convergence to
accepting states.

\subsection{Automaton Update\label{subsec:Automaton}}

Since the environment is only partially known, this section describes
how the real-time information sensed by the robot during the runtime
can be used to update the system model to facilitate motion planning.
The robot starts with an initial, possibly imprecise, knowledge about
the environment. A potential cause of infeasible task specifications
is the imprecise state labels. Due to limited local sensing capability,
let $Q_{N}$ denote the set of sensible neighboring states and le\textcolor{black}{t
$\left\llbracket s_{\mathcal{P}}\right\rrbracket =\left\{ s_{\mathcal{P}}=\left(q,s_{h},s_{s}\right)\left|q\in Q_{N}\right.\right\} $
denote a class of $s_{\mathcal{P}}$ sharing the same neighboring
states. Specifically, let $\Info\left(s_{\mathcal{P}}\right)=\left\{ L_{\mathcal{P}}\left(s_{\mathcal{P}}^{\prime}\right)\left|s_{\mathcal{P}}^{\prime}\in\Sense\left(s_{\mathcal{P}}\right)\right.\right\} $
denote the newly observed labels of $s_{\mathcal{P}}^{\prime}$ that
are different from the current knowledge, where $\Sense\left(s_{\mathcal{P}}\right)$
represents a local set of states that can be sensed by the robot at
$s_{\mathcal{P}}$. If the sensed labels $L_{\mathcal{P}}\left(s_{\mathcal{P}}^{\prime}\right)$
are consistent with the current knowledge of }\textbf{\textcolor{black}{$s_{\mathcal{P}}^{\prime}$}}\textcolor{black}{,
$\Info\left(s_{\mathcal{P}}\right)=\emptyset$. Otherwise, the properties
of $s_{\mathcal{P}}^{\prime}$ need to be updated.}

Let $\boldsymbol{J\left(\left\llbracket s_{\mathcal{P}}\right\rrbracket \right)}\in\mathbb{R}^{\left|\left\llbracket s_{\mathcal{P}}\right\rrbracket \right|}$
denote the stacked $J$ for all $s_{\mathcal{P}}\in\left\llbracket s_{\mathcal{P}}\right\rrbracket $.
For $i,j=1,\ldots\left|S_{\mathcal{P}}\right|$, let $\mathrm{\mathbf{H}_{\mathcal{P}}}\in\mathbb{R}^{\left|S_{\mathcal{P}}\right|\times\left|S_{\mathcal{P}}\right|}$
denote a matrix where the $\left(i,j\right)$th entry of $\mathrm{\mathbf{H}_{\mathcal{P}}}$
represents $\mathrm{h_{\mathcal{P}}}\left(s_{i}^{\mathcal{P}},s_{j}^{\mathcal{P}}\right)$
and let $\mathrm{\mathbf{V}_{\mathcal{P}}}\in\mathbb{R}^{\left|S_{\mathcal{P}}\right|\times\left|S_{\mathcal{P}}\right|}$
denote a matrix where the $\left(i,j\right)$th entry of $\mathrm{\mathbf{V}_{\mathcal{P}}}$
represents the violation cost $\mathrm{v_{\mathcal{P}}}\left(s_{i}^{\mathcal{P}},s_{j}^{\mathcal{P}}\right)$.
The terms $\boldsymbol{J}$ , $\mathrm{\mathbf{H}_{\mathcal{P}}}$
and $\mathrm{\mathbf{V}_{\mathcal{P}}}$ are initialized from the
initial knowledge of the environment. Algorithm \ref{Alg1} outlines
how $\mathrm{\mathbf{H}_{\mathcal{P}}}$, $\mathrm{\mathbf{V}_{\mathcal{P}}}$
and $\boldsymbol{J}$ are updated based on the locally sensed information
to facilitate motion planning in line 2-9. At each step, if $\Info\left(s_{\mathcal{P}}\right)\neq\emptyset$,
the energy function $\boldsymbol{J}$ for each states of $\left\llbracket s_{\mathcal{P}}\right\rrbracket $
is updated by Algorithm \ref{Alg1}.

\begin{algorithm}
\caption{ \label{Alg1}Automaton Update}

\scriptsize

\singlespacing

\begin{algorithmic}[1]

\Procedure {Input: } { the current state $s_{\mathcal{P}}=\left({q,s_{h},s_{s}}\right)$,	the current $J\left(\left\llbracket s_{\mathcal{P}}\right\rrbracket \right)$, $\mathcal{F}^{*}$, and $\Info\left(s_{\mathcal{P}}\right)$ }

{Output: } { the updated $\boldsymbol{J}'$ }

\If { $\Info\left(s_{\mathcal{P}}\right)\neq\emptyset$}

\For { all $s_{\mathcal{P}}^{\prime}=\left(q',s_{h}',s_{s}'\right)\in\Sense\left(s_{\mathcal{P}}\right)$ such that $L_{\mathcal{P}}\left(s_{\mathcal{P}}^{\prime}\right)\in\Info\left(s_{\mathcal{P}}\right)$
}

\For { all $\hat{s}_{\mathcal{P}}^{\prime}$ such that $(s_{\mathcal{P}}^{\prime},\hat{s}_{\mathcal{P}}^{\prime})\in\varDelta_{\mathcal{\mathcal{P}}}$
}

\State Update the labels of $L_{\mathcal{P}}\left(s'_{\mathcal{P}}\right)$
according to $L\left(q'\right)$

\State Update $\mathrm{\mathbf{H}_{\mathcal{P}}}$ and $\mathrm{\mathbf{V}_{\mathcal{P}}}$ 

\EndFor

\EndFor

\State Update $J\left(\left\llbracket s_{\mathcal{P}}\right\rrbracket \right)$
based on (\ref{eq:energy})

\EndIf

\EndProcedure

\end{algorithmic}
\end{algorithm}

\begin{lem}
	\textcolor{black}{\label{Lemma:F_star}The largest self-reachable
		set $\mathcal{F}^{*}$ remains the same during the automaton update
		in Algorithm \ref{Alg1}.}
\end{lem}
\begin{IEEEproof}
	\textcolor{black}{By neglecting the cost of transitions in Section
		\ref{subsec:Energy}, the relaxed product automaton $\mathcal{P}_{\left(S_{\mathcal{P}},\varDelta_{\mathcal{P}}\right)}$
		can be treated as a directed graph $\mathcal{G}_{\left(S_{\mathcal{P}},\varDelta_{\mathcal{P}}\right)}.$
		By Def.\ref{def:RPA}, Alg. \ref{Alg1} only updates the cost of each
		transition. As a result, the topological structure of $\mathcal{G}_{\left(S_{\mathcal{P}},\varDelta_{\mathcal{P}}\right)}$
		and its corresponding $\mathcal{F}^{*}$ remain the same.}
\end{IEEEproof}
\begin{rem}
	The construction of $\mathcal{F}^{*}$ in \cite{Ding2014} involves
	the computation of $d\left(s_{\mathcal{P}},s_{\mathcal{P}}^{\prime}\right)$
	for all $s_{\mathcal{P}}^{\prime}\in\mathcal{F}_{\mathcal{P}}$ and
	the check of terminal conditions, leading to the computational complexity
	of $O\left(\left|\mathcal{F}_{\mathcal{P}}\right|^{3}+\left|S_{\mathcal{P}}\right|^{2}\times\left|\mathcal{F}_{\mathcal{P}}\right|^{2}+\left|\mathcal{F}_{\mathcal{P}}\right|\right)$.
	In contrast, Lemma \ref{Lemma:F_star} indicates that $\mathcal{F}^{*}$
	in this work only needs to be updated whenever newly sensed information
	different from its knowledge is obtained, which reduces the complexity.
	In the worst case, the complexity is $\left|Q_{N}\right|$. Instead
	of computing the whole relaxed product automaton, Algorithm \ref{Alg1}
	only updates partial information of the systems.
\end{rem}

\section{Control Synthesis of LTL Motion Planning}

This section presents a RHC-based online motion planning strategy
that optimizes accumulated utilities over a predefined finite horizon
subject to energy function based constraints, where the accumulated
utilities take into account both the time-varying reward and the violation
cost, while the energy function based constraints enforce the satisfaction
of the acceptance condition of the relaxed product automaton $\mathcal{P}$.

\subsection{Receding Horizon Control}

The general idea of RHC is to generate a predicted optimal trajectory
at each time step by solving an online optimization problem to maximize
a utility function over a finite horizon $N$. With only the first
predicted step applied, the optimization problem is repeatedly solved
to predict optimal trajectories. Specifically, based on the current
state $s_{k}^{\mathcal{P}}$, let \textbf{$\bar{\boldsymbol{s}}_{k}^{\mathcal{P}}=s_{1\mid k}^{\mathcal{P}}s_{2\mid k}^{\mathcal{P}}\ldots s_{N\mid k}^{\mathcal{P}}$}
denote a predicted trajectory of\textbf{ }horizon $N$ at time $k$
from $s_{k}^{\mathcal{P}}$, where the $i$th predicted state $s_{i\mid k}^{\mathcal{P}}\in S_{\mathcal{P}}$
satisfies $\left(s_{i\mid k}^{\mathcal{P}},s_{i+1\mid k}^{\mathcal{P}}\right)\in\varDelta_{\mathcal{\mathcal{P}}}$
for all $i=1,\ldots,N-1$, and $\left(s_{k}^{\mathcal{P}},s_{1\mid k}^{\mathcal{P}}\right)\in\varDelta_{\mathcal{\mathcal{P}}}$.
Let $\Path\left(s_{k}^{\mathcal{P}},N\right)$ be the set of trajectories
of horizon $N$ generated from $s_{k}^{\mathcal{P}}$. Note that a
predicted trajectory $\bar{\boldsymbol{s}}_{k}^{\mathcal{P}}\in\Path\left(s_{k}^{\mathcal{P}},N\right)$
can uniquely project to a path $\gamma_{\mathcal{T}}\left(\bar{\boldsymbol{s}}_{k}^{\mathcal{P}}\right)=\boldsymbol{q}=q_{1}\cdots q_{N}$
on $\mathcal{T}$, where $\gamma_{\mathcal{T}}\left(s_{i\mid k}^{\mathcal{P}}\right)=q_{i}$,
$\forall i=1,\ldots,N$.

\textcolor{black}{The finite horizon $N$ is selected based on the
	robot's local sensing such that the labels $L_{\mathcal{P}}\left(q_{i}\right)$
	and the reward $R_{k}\left(q_{i}\right)$, $\forall i=1,\ldots,N$,
	are all observable by the robot at time $k$. The accumulated reward
	along the predicted trajectory $\bar{\boldsymbol{s}}_{k}^{\mathcal{P}}$
	is $\mathbf{R}\left(\gamma_{\mathcal{T}}\left(\bar{\boldsymbol{s}}_{k}^{\mathcal{P}}\right)\right)=\stackrel[i=1]{N}{\sum}R_{k}\left(\gamma_{\mathcal{T}}\left(s_{i\mid k}^{\mathcal{P}}\right)\right).$}

Once a predicted step $k$ of RHC is implemented, the hard and soft
violation cost induced from the current state $s_{k}^{\mathcal{P}}$
to the next predicted step $s_{1\mid k}^{\mathcal{P}}$ are considered,
i.e., $h_{\mathcal{P}}\left(s_{k}^{\mathcal{P}},s_{1\mid k}^{\mathcal{P}}\right)$
and $\mathbf{V}\left(s_{k}^{\mathcal{P}}\right)=\beta\cdot\mathrm{v_{\mathcal{P}}}\left(s_{k}^{\mathcal{P}},s_{1\mid k}^{\mathcal{P}}\right).$
The utility function of RHC is then designed as
\begin{equation}
\mathbf{U}\left(\bar{\boldsymbol{s}}_{k}^{\mathcal{P}}\right)=-h_{\mathcal{P}}\left(s_{k}^{\mathcal{P}},s_{1\mid k}^{\mathcal{P}}\right)+\mathbf{R}\left(\gamma_{\mathcal{T}}\left(\bar{\boldsymbol{s}}_{k}^{\mathcal{P}}\right)\right)\min\left\{ e^{-\kappa\mathbf{V}\left(s_{k}^{\mathcal{P}}\right)},1\right\} \label{eq:utility}
\end{equation}
where $\kappa\in\mathbb{R}^{+}$ is a tuning parameter indicating
how aggressively a predicted path is penalized by violating the soft
task constraints and the non-zero violation $\mathbf{V}\left(s_{k}^{\mathcal{P}}\right)$
in (\ref{eq:utility}) would enforce the decrease of $\mathbf{U}\left(\bar{\boldsymbol{s}}_{k}^{\mathcal{P}}\right)$.
If $h_{\mathcal{P}}\left(s_{k}^{\mathcal{P}},s_{1\mid k}^{\mathcal{P}}\right)=\infty$,
it indicates that $\mathbf{U}\left(\bar{\boldsymbol{s}}_{k}^{\mathcal{P}}\right)$
is negative infinite. By applying a larger $\kappa$ optimizing $\mathbf{U}\left(\bar{\boldsymbol{s}}_{k}^{\mathcal{P}}\right)$
tends to bias the selection of paths towards the objectives, in the
decreasing order, of 1) hard task $\phi_{h}$ satisfaction, 2) fulfilling
soft task $\phi_{h}$ as much as possible, and 3) time-varying rewards
locally optimization.

Since maximizing $\mathbf{U}\left(\bar{\boldsymbol{s}}_{k}^{\mathcal{P}}\right)$
alone cannot guarantee the satisfaction of the acceptance condition
of $\mathcal{P}$, energy function the energy function based constraints
are incorporated. We first select initial states from $S_{\mathcal{P}0}$
that can reach the set $\mathcal{F}^{*}$. The RHC executing on $S_{\mathcal{P}0}$
is designed as

\begin{equation}
\begin{aligned}\boldsymbol{\bar{s}}_{0,\text{opt}}^{\mathcal{P}}= & \underset{\bar{\boldsymbol{s}}_{0}^{\mathcal{P}}\in\Path\left(s_{0}^{\mathcal{P}},N\right)}{\arg\max}\mathbf{U}\left(\bar{\boldsymbol{s}}_{0}^{\mathcal{P}}\right)\\
& \text{ subject to : }J\left(s_{0}^{\mathcal{P}}\right)<\infty.
\end{aligned}
\label{eq:optimization_initial}
\end{equation}
The constraint $J\left(s_{0}^{\mathcal{P}}\right)<\infty$ in (\ref{eq:optimization_initial})
is critical, since a bounded energy $J\left(s_{0}^{\mathcal{P}}\right)$
guarantees the existence of a satisfying trajectory from $s_{0}^{\mathcal{P}}$
over $\mathcal{P}$. According to the working principle of RHC, the
first element of the optimal trajectory $\bar{\boldsymbol{s}}^{\mathcal{P}*}$
can be determined as $s_{0}^{\mathcal{P}*}=s_{1\mid0,\text{opt}}^{\mathcal{P}},$
where $s_{1\mid0,\text{opt}}^{\mathcal{P}}$ is the first element
of $\boldsymbol{\bar{s}}_{0,\text{opt}}^{\mathcal{P}}$ obtained from
(\ref{eq:optimization_initial}).

After determining the initial state $s_{0}^{\mathcal{P}*}$, RHC will
be employed repeatedly to determine the optimal states $s_{k}^{\mathcal{P}*}$
for $\mathit{k}=1,2,\ldots$. At each time instant $k$, a predicted
optimal trajectory $\bar{\boldsymbol{s}}_{k,\text{opt}}^{\mathcal{P}}=s_{1\mid k,\text{opt}}^{\mathcal{P}}s_{2\mid k,\text{opt}}^{\mathcal{P}}\ldots s_{N\mid k,\text{opt}}^{\mathcal{P}}$
will be constructed based on $s_{k-1}^{\mathcal{P}*}$ and $\bar{\boldsymbol{s}}_{k-1,\text{opt}}^{\mathcal{P}}$
obtained at the previous time $k-1$. Note that only $s_{1\mid k,\text{opt}}^{\mathcal{P}}$
will be applied at time $k$, i.e., $\text{\ensuremath{s_{k}^{\mathcal{P}*}}}=s_{1\mid k,\text{opt}}^{\mathcal{P}}$,
which will then be used with $\boldsymbol{\bar{s}}_{k,\text{opt}}^{\mathcal{P}}$
to generate $\bar{\boldsymbol{s}}_{k+1,\text{opt}}^{\mathcal{P}}.$
\begin{thm}
	\label{thm:RHC}For each time $k=1,2\ldots$, provided $s_{k-1}^{\mathcal{P}*}$
	and $\bar{\boldsymbol{s}}_{k-1,\text{opt}}^{\mathcal{P}}$ from previous
	time step, consider a receding horizon control (RHC)
	\begin{equation}
	\bar{\boldsymbol{s}}_{k,\text{opt}}^{\mathcal{P}}=\underset{\boldsymbol{\bar{s}}_{k}^{\mathcal{P}}\in\Path\left(s_{k-1}^{\mathcal{P}*},N\right)}{\arg\max}\mathbf{U}\left(\boldsymbol{\bar{s}}_{k}^{\mathcal{P}}\right)\label{eq:RHC}
	\end{equation}
	subject to the following constraints\textup{:}
	\begin{enumerate}
		\item \textup{$J\left(s_{N\mid k}^{\mathcal{P}}\right)<J\left(s_{N\mid k-1,\text{opt}}^{\mathcal{P}}\right)$
		}if \textup{$J\left(s_{k-1}^{\mathcal{P}*}\right)>0$ }and \textup{$J\left(s_{i\mid k-1,\text{opt}}^{\mathcal{P}}\right)\neq0$
		}for all\textup{ $i=1,\ldots,N$;}.
		\item $J\left(s_{i_{0}\left(\boldsymbol{s}_{k-1,\text{opt}}^{\mathcal{P}}\right)-1\mid k}^{\mathcal{P}}\right)=0$
		if $J\left(s_{k-1}^{\mathcal{P}*}\right)>0$ and $J\left(s_{i\mid k-1,\text{opt}}^{\mathcal{P}}\right)=0$
		for some \textup{$i=1,\ldots,N$;}
		\item $J\left(s_{N\mid k}^{\mathcal{P}}\right)<\infty$ if $J\left(s_{k-1}^{\mathcal{P}*}\right)=0$.
	\end{enumerate}
	Applying $s_{k}^{\mathcal{P}*}=s_{1\mid k,\text{opt}}^{\mathcal{P}}$
	at each time $k$, the optimal trajectory $\boldsymbol{\bar{s}}^{\mathcal{P}*}=s_{0}^{\mathcal{P}*}s_{1}^{\mathcal{P}*}\ldots$
	is guaranteed to satisfy the acceptance condition of $\mathcal{P}$.
\end{thm}
\begin{IEEEproof}
	Consider a state $s_{k-1}^{\mathcal{P}*}\in S_{\mathcal{P}},$ $\forall k=1,2,\ldots$,
	and $\Path\left(s_{k-1}^{\mathcal{P}*},N\right)$ represents the set
	of all possible paths starting from $s_{k-1}^{\mathcal{P}*}$ with
	horizon $N$. Since not all predicated trajectories maximizing the
	utility $\mathbf{U}\left(\boldsymbol{\bar{s}}_{k}^{\mathcal{P}}\right)$,
	$\bar{\boldsymbol{s}}_{k}^{\mathcal{P}}\in\Path\left(s_{k-1}^{\mathcal{P}*},N\right)$,
	in (\ref{eq:RHC}) are guaranteed to be accepting by $\mathcal{P}$,
	additional constraints need to be imposed. Note that the energy function
	$J\left(s_{k-1}^{\mathcal{P}*}\right)$ defined in (\ref{eq:energy})
	indicates the distance from the current state $s_{k-1}^{\mathcal{P}*}$
	to $\mathcal{F}^{*}.$ A trajectory on $\mathcal{P}$ is accepting
	if the trajectory can intersect $\mathcal{F}^{*}$ infinitely many
	times. Therefore, the key idea of the design of the constraints for
	(\ref{eq:RHC}) is to ensure the energy of the states along the trajectory
	eventually decreases to zero. Following this idea, different cases
	are considered.
	
	(i) Case 1: If $J\left(s_{k-1}^{\mathcal{P}*}\right)>0$ and $J\left(s_{i\mid k-1,\text{opt}}^{\mathcal{P}}\right)\neq0$
	for all $i=1,\ldots,N$, the constraint $J\left(s_{N\mid k}^{\mathcal{P}}\right)<J\left(s_{N\mid k-1,\text{opt}}^{\mathcal{P}}\right)$
	is enforced. Recall that $\boldsymbol{\bar{s}}_{k-1,\text{opt}}^{\mathcal{P}}=s_{1\mid k-1,\text{opt}}^{\mathcal{P}}s_{2\mid k-1,\text{opt}}^{\mathcal{P}}\ldots s_{N\mid k-1,\text{opt}}^{\mathcal{P}}$
	is the predicted optimal trajectory at the previous time $k-1$. The
	energy $J\left(s_{k-1}^{\mathcal{P}*}\right)>0$ indicates that there
	exists a trajectory from $s_{k-1}^{\mathcal{P}*}$ to $\mathcal{F}^{*}$,
	and $J\left(s_{i\mid k-1,\text{opt}}^{\mathcal{P}}\right)\neq0$ for
	all $i=1,\ldots,N$ indicates $\boldsymbol{\bar{s}}_{k-1,\text{opt}}^{\mathcal{P}}$
	does not intersect $\mathcal{F}^{*}$. The constraint $J\left(s_{N\mid k}^{\mathcal{P}}\right)<J\left(s_{N\mid k-1,\text{opt}}^{\mathcal{P}}\right)$
	enforces that the energy of the last state $s_{N\mid k}^{\mathcal{P}}$
	in the predicted trajectory at the current time $k$ must be less
	than that of the previously predicted $\boldsymbol{\bar{s}}_{k-1,\text{opt}}^{\mathcal{P}}$,
	which indicates the energy along $\boldsymbol{\bar{s}}_{k,\text{opt}}^{\mathcal{P}}$
	strictly decreases at each iteration $k$. Note that, based on Theorem
	\ref{Thm:energy}, there always exists a state $s_{\mathcal{P}}^{\prime}$
	on $\mathcal{P}$ satisfying $\left(s_{N\mid k-1,\text{opt}}^{\mathcal{P}},s_{\mathcal{P}}^{\prime}\right)\in\varDelta_{\mathcal{P}}$
	and $J\left(s_{\mathcal{P}}^{\prime}\right)<J\left(s_{N\mid k-1,\text{opt}}^{\mathcal{P}}\right)$.
	Therefore, if we can construct a trajectory $\boldsymbol{\bar{s}}_{k}^{\mathcal{P}}=s_{1\mid k}^{\mathcal{P}},\ldots,s_{N\mid k}^{\mathcal{P}}$
	with $s_{i\mid k}^{\mathcal{P}}=s_{i+1\mid k-1,\text{opt}}^{\mathcal{P}}$
	and $s_{N\mid k,\text{opt}}^{\mathcal{P}}=s_{\mathcal{P}}^{\prime}$
	for all $i=1,\ldots,N-1$, the problem (\ref{eq:RHC}) is guaranteed
	to have at least one solution for Case 1.
	
	(ii) Case 2: If $J\left(s_{i\mid k-1,\text{opt}}^{\mathcal{P}}\right)=0$
	for some $i=1,\ldots,N$, $\boldsymbol{\bar{s}}_{k-1,\text{opt}}^{\mathcal{P}}$
	intersects $\mathcal{F}^{*}$. Let $i_{0}\left(\boldsymbol{\bar{s}}_{k-1,\text{opt}}^{\mathcal{P}}\right)$
	be the index of the first occurrence in $\bar{\boldsymbol{s}}_{k-1,\text{opt}}^{\mathcal{P}}$
	where $J\left(s_{i_{0}\mid k-1}^{\mathcal{P}}\right)=0$. The constraint
	$J\left(s_{i_{0}\left(\boldsymbol{s}_{k-1}^{\mathcal{P}}\right)-1\mid k}^{\mathcal{P}}\right)=0$
	enforces the predicted trajectory at the current time $k$ to have
	energy 0 (i.e., intersect $\mathcal{F}^{*}$), if the previously predicted
	trajectory $\boldsymbol{\bar{s}}_{k-1,\text{opt}}^{\mathcal{P}}$
	does so. To show that the problem (\ref{eq:RHC}) has at least one
	solution for Case 2, we can always construct $\boldsymbol{\bar{s}}_{k}^{\mathcal{P}}=s_{1\mid k}^{\mathcal{P}},\ldots,s_{N\mid k}^{\mathcal{P}}$
	by letting $s_{i\mid k}^{\mathcal{P}}=s_{i+1\mid k-1,\text{opt}}^{\mathcal{P}}$
	and $s_{N\mid k}^{\mathcal{P}}=s_{\mathcal{P}}^{\prime}$ for all
	$i=1,\ldots,N-1$, where $s_{\mathcal{P}}^{\prime}$ can be any state
	on $\mathcal{P}$ satisfying $\left(s_{N\mid k-1,\text{opt}}^{\mathcal{P}},s_{\mathcal{P}}^{\prime}\right)\in\varDelta_{\mathcal{P}}$
	and $J\left(s_{\mathcal{P}}^{\prime}\right)<\infty$.
	
	(iii) Case 3: If $J\left(s_{k-1}^{\mathcal{P}*}\right)=0,$ it indicates
	$s_{k-1}^{\mathcal{P}*}\in\mathcal{F}^{*}$. The constraint $J\left(s_{N\mid k}^{\mathcal{P}}\right)<\infty$
	only requires the predicted trajectory $\boldsymbol{\bar{s}}_{k}^{\mathcal{P}}$
	ending at a state with bounded energy, where Cases 1 and 2 can then
	be applied to enforce the following sequence $s_{k+1}^{\mathcal{P}*}s_{k+2}^{\mathcal{P}*}\ldots$
	converging to $\mathcal{F}^{*}$. To show that there always exists
	$s_{N\mid k}^{\mathcal{P}}$ with $J\left(s_{N\mid k}^{\mathcal{P}}\right)<\infty$,
	note that there exists a state $s_{\mathcal{P}}^{\prime}$ satisfying
	$\left(s_{k-1}^{\mathcal{P}*},s_{\mathcal{P}}^{\prime}\right)\in\varDelta_{\mathcal{P}}$
	and $J\left(s_{\mathcal{P}}^{\prime}\right)<\infty$. Let $s_{1\mid k}^{\mathcal{P}}=s_{\mathcal{P}}^{\prime}$.
	Based on Theorem \ref{Thm:energy}, we can always construct $\bar{\boldsymbol{s}}_{k}^{\mathcal{P}}=s_{1\mid k}^{\mathcal{P}},\ldots,s_{N\mid k}^{\mathcal{P}}$
	such that $J\left(s_{i\mid k}^{\mathcal{P}}\right)<J\left(s_{i+1\mid k}^{\mathcal{P}}\right)$
	for all $i=1,\ldots,N-1$, and $J\left(s_{N\mid k}^{\mathcal{P}}\right)<\infty$.
	Consequently, the problem (\ref{eq:RHC}) has solutions for Case 3.
\end{IEEEproof}
\textcolor{black}{Analogous to the analysis in \cite{Ding2014}, the
	energy function based constraints (\ref{eq:RHC}) in Theorem \ref{thm:RHC}
	ensure that $\boldsymbol{\bar{s}}^{\mathcal{P}*}=s_{0}^{\mathcal{P}*}s_{1}^{\mathcal{P}*}\ldots$.
	intersects the accepting states $\mathcal{F}_{\mathcal{P}}$ infinitely,
	resulting in the satisfaction of the acceptance condition of $\mathcal{P}$.
	If the RHC yields an negative infinite utility, the hard constraint
	is violated and the robot fails to accomplish the task. In this paper,
	we assume $\phi_{h}$ is always feasible.}

\subsection{Control Synthesis}

The control synthesis of the LTL online motion planning strategy is
outlined in Algorithm \ref{Alg2}. In Lines 1-3, an off-line computation
is first performed over $\mathcal{P}$ to obtain an initial $\boldsymbol{J}$
and an initial violation cost $\mathrm{\mathbf{V}_{\mathcal{P}}}$.
At time $k=0$, the receding horizon control (\ref{eq:optimization_initial})
is applied to determine $s_{0}^{\mathcal{P}*}$ in Lines 4-7. Due
to the dynamic and uncertain nature of the environment, Algorithm
\ref{Alg1} is applied at each time $k>0$ to update $\boldsymbol{J\left(\left\llbracket s_{\mathcal{P}}\right\rrbracket \right)}$
and $\mathrm{\mathbf{V}_{\mathcal{P}}}$ based on local sensing in
Lines 9-10. The RHC (\ref{eq:RHC}) is then employed based on the
previously determined $s_{k-1}^{\mathcal{P}*}$ to generate $\boldsymbol{\bar{s}}_{k,\text{opt}}^{\mathcal{P}}$,
where the next state is determined as $s_{k}^{\mathcal{P}*}=s_{1\mid k,\text{opt}}^{\mathcal{P}}$
in Lines 11-12. The transition from $s_{k-1}^{\mathcal{P}*}$ to $s_{k}^{\mathcal{P}*}$
is then immediately applied on $\mathcal{P}$, which corresponds to
the movement of the robot at time $k$ from $\gamma_{\mathcal{T}}\left(s_{k-1}^{\mathcal{P}*}\right)$
to $\gamma_{\mathcal{T}}\left(s_{k}^{\mathcal{P}*}\right)$ on $\mathcal{T}$
in Line 13. Repeating the process can generate a trajectory $\boldsymbol{\bar{s}}^{\mathcal{P}*}=s_{0}^{\mathcal{P}*}s_{1}^{\mathcal{P}*}\ldots$
that optimizes the utilities while satisfying the acceptance condition
of $\mathcal{P}$. If $J\left(s_{0}^{\mathcal{P}}\right)=\infty$,
there exists no trajectory that satisfies $\phi_{h}$ in Line 17.
\begin{thm}[Correctness of Algorithm \ref{Alg2}]
	\label{thm:Proof_Alg2} Given a weighted DTS $\mathcal{T}=\left\{ Q,q_{0},\delta,\Pi,L,\omega\right\} $
	and $\mathcal{B}_{h}$ and $\mathcal{B}_{s}$ corresponding to $\phi_{h}$
	and $\phi_{s}$, respectively, if there exists an initial state $s_{0}^{\mathcal{P}}\in S_{\mathcal{P}0}$
	with $J\left(s_{0}^{\mathcal{P}}\right)<\infty$, the trajectory generated
	by Algorithm \ref{Alg2} is guaranteed to satisfy the acceptance condition
	of $\mathcal{P}$. 
\end{thm}
\begin{IEEEproof}
	The existence of an initial state $s_{0}^{\mathcal{P}}\in S_{\mathcal{P}0}$
	with $J\left(s_{0}^{\mathcal{P}}\right)<\infty$ indicates the existence
	of a solution to (\ref{eq:optimization_initial}). The solution $\boldsymbol{\bar{s}}_{0,\text{opt}}^{\mathcal{P}}$
	from (\ref{eq:optimization_initial}) determines the first element
	of the trajectory $\bar{\boldsymbol{s}}^{\mathcal{P}*}$, i.e., $s_{0}^{\mathcal{P}*}=s_{1\mid0,\text{opt}}^{\mathcal{P}}$,
	with $J\left(s_{0}^{\mathcal{P}*}\right)<\infty$, from which (\ref{eq:RHC})
	can be applied recursively to determine the rest elements $s_{k}^{\mathcal{P}*}$,
	$k=1,\ldots,$ of $\boldsymbol{\bar{s}}^{\mathcal{P}*}$. Particularly,
	for each time $k$, if the constraint 1 in Theorem \ref{thm:RHC}
	is satisfied, $J\left(s_{k-1}^{\mathcal{P}*}\right)<\infty$ indicates
	there exist other states with lower energy by Theorem \ref{Thm:energy}.
	Hence, repeatedly applying (\ref{eq:RHC}) can generate a set of predicted
	optimal paths with $J\left(s_{N\mid k}^{\mathcal{P}}\right)>J\left(s_{N\mid k+1}^{\mathcal{P}}\right)>J\left(s_{N\mid k+2}^{\mathcal{P}}\right)\ldots$
	satisfying $J\left(s_{N\mid j}^{\mathcal{P}}\right)=0$ for some $j>k$.
	If the constraints 2) and 3) in Theorem \ref{thm:RHC} are satisfied,
	the predicted optimal trajectories $\boldsymbol{\bar{s}}_{k,\text{opt}}^{\mathcal{P}}$
	from (\ref{eq:RHC}) will lead $\boldsymbol{\bar{s}}^{\mathcal{P}*}$
	to states with zero energy, which implies the intersections with $\mathcal{F}^{*}$.
	Repeating the process described above, the resulting trajectory $\boldsymbol{\bar{s}}^{\mathcal{P}*}$
	from Algorithm \ref{Alg2} satisfies the acceptance condition.\textcolor{black}{{}
		Moreover, if $J\left(s_{k}^{\mathcal{P}}\right)<\infty$, it indicates
		there exists a run satisfying $\phi_{h}$ and the violation cost $-h_{\mathcal{P}}\left(s_{k}^{\mathcal{P}},s_{1\mid k}^{\mathcal{P}}\right)$
		in RHC ensures the satisfaction of $\phi_{h}$ since only the first
		predicted step is applied.}
\end{IEEEproof}
\begin{cor}
	\label{cor: feasible correctness }Given a weighted DTS $\mathcal{T}=\left\{ Q,q_{0},\delta,\Pi,L,\omega\right\} $,
	$\mathcal{B}_{h}$ and $\mathcal{B}_{s}$, if $\varphi_{s}$ is feasible,
	the solution of Algorithm \ref{Alg2} fully satisfies the task $\varphi=\varphi_{h}\wedge\varphi_{s}$
	exactly with $\kappa$ in (\ref{eq:utility}) selected sufficiently
	large.
\end{cor}
Since Corollary \ref{cor: feasible correctness } is an immediate
result of Theorem \ref{Thm:hard_components} and Theorem \ref{thm:Proof_Alg2},
its proof is omitted. 

\begin{algorithm}
	\caption{\label{Alg2}Control synthesis of LTL online motion planning}
	
	\scriptsize
	
	\singlespacing
	
	\begin{algorithmic}[1]
		
		\Procedure {Input:} {The DTS $\mathcal{T}=\left\{ Q,q_{0},\delta,\Pi,L,\omega\right\} $
			and the NBA $\mathcal{B}_{h},\mathcal{B}_{s}$ corresponding to the
			user-specified LTL formula $\phi=\phi_{h}\land\phi_{s}$ }
		
		{Output: } { The trajectory $\boldsymbol{\bar{s}}^{\mathcal{P}*}=s_{0}^{\mathcal{P}*}s_{1}^{\mathcal{P}*}\ldots$
		}
		
		\Statex \textbf{Off-line Execution:}
		
		\State Construct the relaxed product automaton \textcolor{black}{$\mathcal{P}=\mathcal{T}\times\mathcal{B}_{h}\times\mathcal{B}_{s}$}
		
		\State Construct $\mathcal{F}^{*}$, and initialize
			$\mathrm{\mathbf{H}_{\mathcal{P}}}$, $\mathrm{\mathbf{V}_{\mathcal{P}}}$
			and $\boldsymbol{J}$
		
		\Statex \textbf{online Execution:}
		
		\If { $\exists s_{0}^{\mathcal{P}}\in S_{\mathcal{P}0}$
			 $J\left(s_{0}^{\mathcal{P}}\right)<\infty$ }
		
		\State Solve (\ref{eq:optimization_initial}) for $\boldsymbol{\bar{s}}_{0,\text{opt}}^{\mathcal{P}}$
		
		\State $s_{0}^{\mathcal{P}*}=s_{1\mid0,\text{opt}}^{\mathcal{P}}$
		and $k\leftarrow1$
		
		\While {\textbf{ $k>0$}}
		
		\State Apply automaton update at $s_{k-1}^{\mathcal{P}*}$ in Algorithm
		\ref{Alg1} based on local sensing
		
		\State Locally observe rewards $R_{k}\left(\gamma_{\mathcal{T}}\left(s_{k-1}^{\mathcal{P}*}\right)\right)$
		
		\State Solve (\ref{eq:RHC}) for $\boldsymbol{\bar{s}}_{k,\text{opt}}^{\mathcal{P}}$
		
		\State Implement corresponding transitions on $\mathcal{P}$
			and $\mathcal{T}$
		
		\State $s_{k}^{\mathcal{P}*}=s_{1\mid k,\text{opt}}^{\mathcal{P}}$
		and $k++$
		
		\EndWhile
		
		\Else
		
		\State There does not exist an accepting run from initial states;
		
		\EndIf
		
		\EndProcedure
		
	\end{algorithmic}
\end{algorithm}

\subsection{Complexity}

Since the off-line execution involves the computation of $\mathcal{P}$,
$\mathcal{F}^{*}$, the initial $\boldsymbol{J}$, and the initial
$\mathrm{\mathbf{V}_{\mathcal{P}}}$, its complexity is $O\left(\left|\mathcal{F}_{\mathcal{P}}\right|^{3}+\left|S_{\mathcal{P}}\right|^{2}\times\left|\mathcal{F}_{\mathcal{P}}\right|^{2}+\left|\mathcal{F}_{\mathcal{P}}\right|\right)$.
For online execution, Since $\mathcal{F}^{*}$ remains the same during
mission operation, as indicated in Algorithm \ref{Alg1}, the worst
case requires $\left|\left\llbracket s_{\mathcal{P}}\right\rrbracket \right|$
runs of Dijkstra's algorithm. In Algorithm \ref{Alg2}, the selected
horizon $N$ in RHC is crucial to the complexity. Suppose the number
of total transitions between states is $\left|\Delta_{\delta}\right|$.
The complexity of recursive computation at each time step is bounded
by $\left|\Delta_{\delta}\right|^{N}$. Another layer of the complexity
in Algorithm \ref{Alg1} comes from the automaton update and the computation
of energy function computation at each iteration. Suppose the number
of $\Sense\left(s_{\mathcal{P}}\right)$ is bounded by $\left|N_{1}\right|$,
which indicates a maximum $\left|N_{1}\right|\times\left|S_{\mathcal{P}}\right|$
runs is required to update the violation cost $\mathrm{\mathbf{V}_{\mathcal{P}}}$
and the state labels. In addition, updating energy $\boldsymbol{J}$
requires $\left|S_{\mathcal{P}}\right|$ runs of the Dijkstra Algorithm
in each iteration. Therefore, the complexity of Algorithm \ref{Alg1}
is at most $O\left(\left|N_{1}\right|\times\left|S_{\mathcal{P}}\right|+\left|S_{\mathcal{P}}\right|\right)$.
Overall, the maximum complexity of the online portion of RHC is $O\left(\left|N_{1}\right|\times\left|S\right|+\left|S_{\mathcal{P}}\right|+\left|\Delta_{\delta}\right|^{N}\right)$.
\begin{figure*}[t]
	\centering{}\includegraphics[scale=0.8]{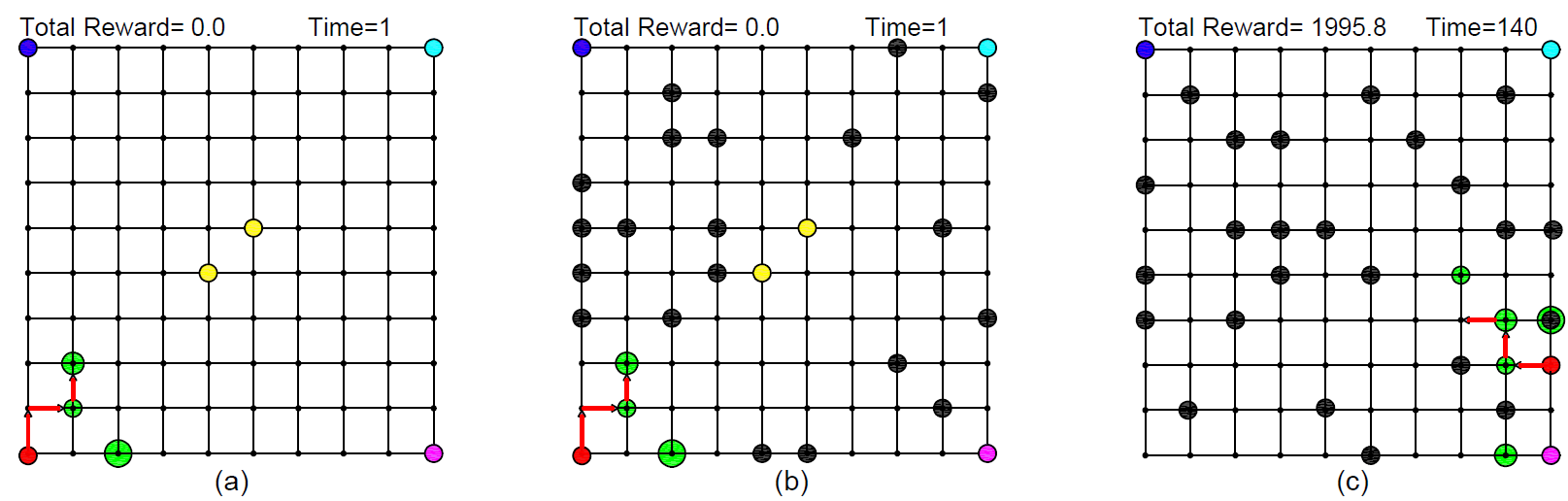}\caption{\label{fig:Snapshots} Snapshots of the environment at different time
		instants. The robot's position is represented by a red circle, while
		the randomly generated rewards within the sensing zone of the vehicle
		are marked by green circles of different sizes proportional to the
		reward value. The red arrow lines represent the predicted trajectory
		at the current time. (a) shows the robot's initial knowledge about
		the environment at $t=1$s. Initially the robot is only informed of
		the positions of $\mathtt{\mathtt{Base}}$, $\mathtt{Supply}$, $\mathtt{Report},$
		and $\mathtt{Survey}$ stations, without any\textit{ a priori} knowledge
		of obstacles (i.e., black circles). (b) shows the real setup of the
		environment scattered with dynamic obstacles at $t=1$s. The environment
		is assumed to be time-varying with yellow circles (i.e., $\mathtt{Survey}$
		stations) on and off at different times, which indicates the environment
		can be infeasible for the robot's desired task. (c) shows that the
		environment is infeasible at $t=140$s, where yellow circles are off.}
\end{figure*}
\begin{figure}
	\centering{}\includegraphics[scale=0.5]{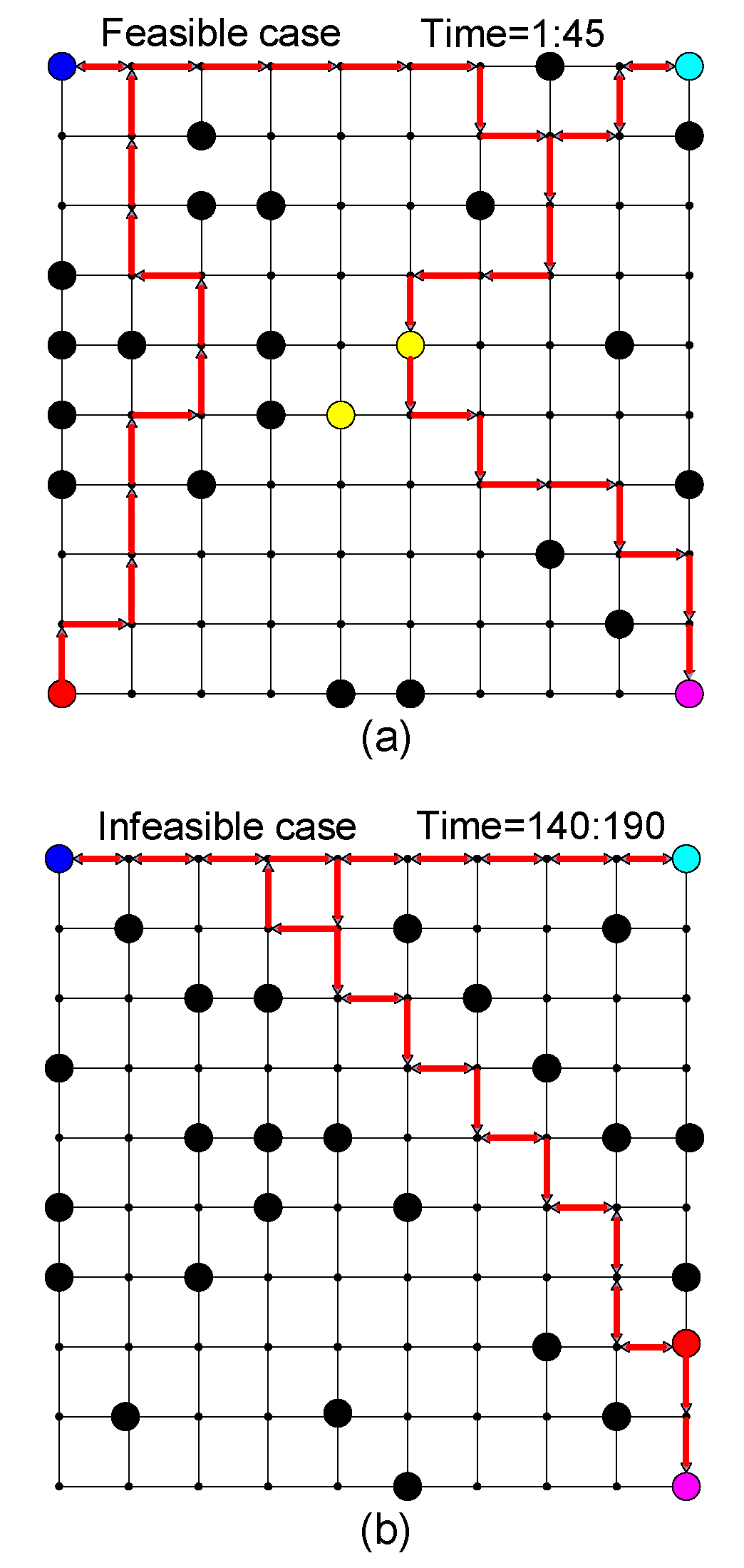}\caption{\label{fig:trajectory} The robot trajectories in feasible and infeasible
		cases of $\phi_{s}$, respectively. In (a), the environment is fully
		feasible from $t=1$s to $t=45$s, and the robot successfully completes
		the desired task (\ref{eq:task1}). In (b), the environment is infeasible
		from $t=140$s to $t=190$s, where yellow circles do not exist. The
		robot revises its motion to only sequentially visit $\mathtt{\mathtt{Base}}$,
		$\mathtt{Supply}$, and $\mathtt{Report}$ stations. In both (a) and
		(b), the planned path maximizes the reward collection in a receding
		horizon manner.}
\end{figure}
\begin{figure}
	\centering{}\includegraphics[scale=0.18]{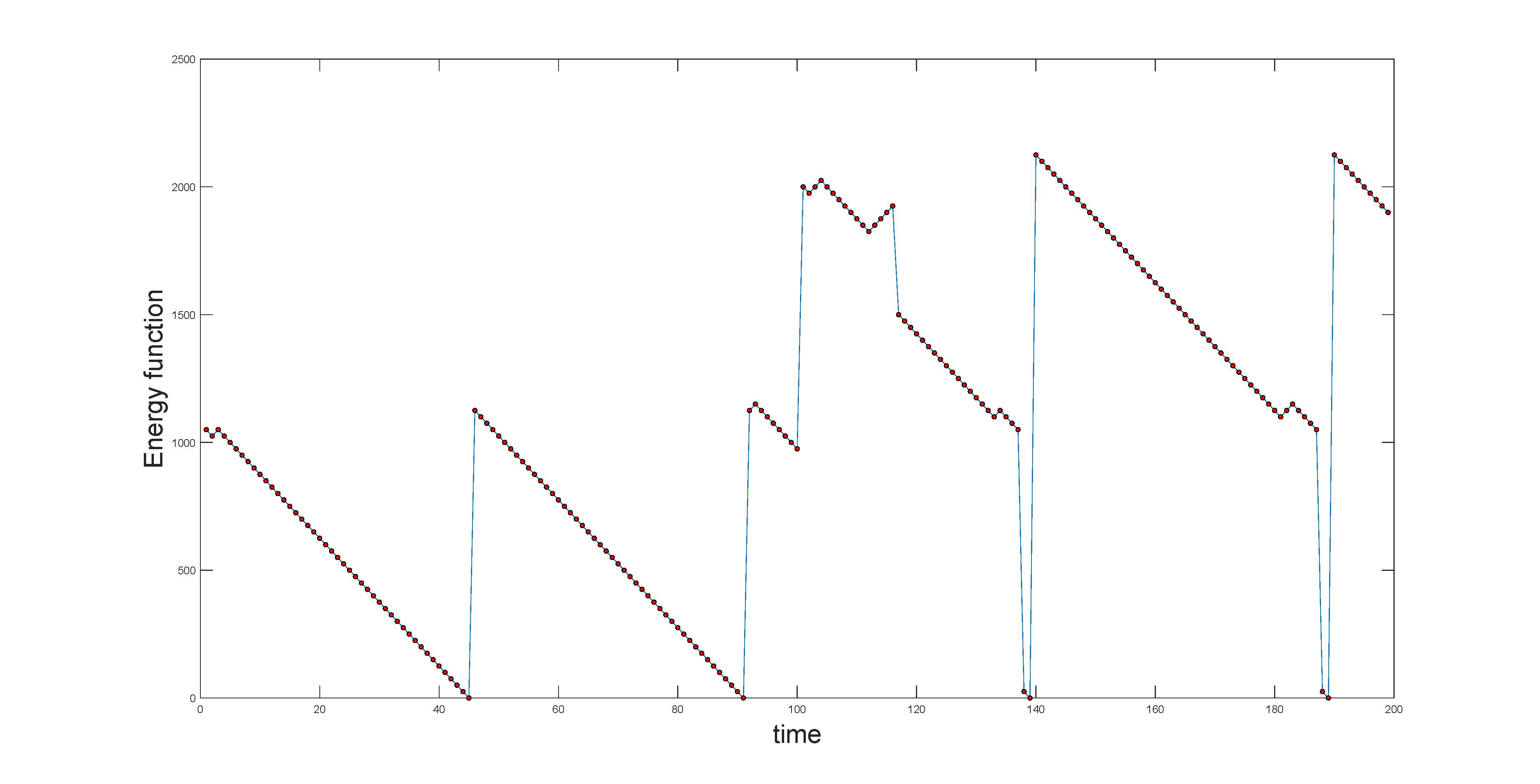}\caption{\label{fig:energy} Plot of the energy function during mission operation.}
\end{figure}
\begin{figure}
	\centering{}\includegraphics[scale=0.4]{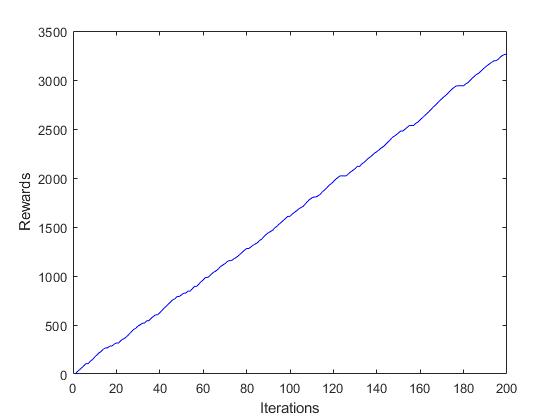}\caption{\label{fig:reward} Plot of accumulative collected time-varying rewards.}
\end{figure}

In the simulation and experiment, we set the penalty parameter ${\beta}=500$
and the tuning parameter $\kappa=100$. The LTL task is $\phi=\phi_{h}\land\phi_{s}$,
where $\phi_{h}=\boxempty\lnot\mathtt{Obstacle}$ and $\phi_{s}$
is defined in Section \ref{subsec:Simu} and \ref{subsec:experiment},
respectively. $\phi_{h}$ was translated to a B\"uchi Automaton $\mathcal{\mathcal{B}}_{h}$
via LTL2BA \cite{Babiak2012} with $\left|S_{h}\right|=1$. 

\section{Case Study}

\subsection{Simulation Results\label{subsec:Simu}}

Consider an application in which a mobile robot performs persistent
surveillance in a dynamic environment. The environment consists of
a $\mathtt{Base}$ station that the robot should visit repeatedly,
$\mathtt{Survey}$ points that indicate the areas of interest that
the robot should explore, a $\mathtt{Report}$ station where the robot
should report its findings after visiting $\mathtt{Survey}$, a $\mathtt{Supply}$
station where the robot can get refueled, and a set of $\mathtt{Obstacle}$
that the robot should avoid during the mission. It is assumed that
the locations of $\mathtt{Supply}$, $\mathtt{Base}$, and $\mathtt{Report}$
are fixed and known to the robot, while the $\mathtt{Survey}$ points
(i.e., the events of interest) are dynamic and can even be occasionally
infeasible for the robot to explore. In addition, the potential $\mathtt{Obstacle}$
are dynamic. The task of the robot is formulated based on LTL as 
\begin{equation}
\begin{aligned}\phi_{s}= & \boxempty\diamondsuit\mathtt{Base}\\
 & \land\boxempty\left(\mathtt{Base}\rightarrow\ocircle\left(\lnot\mathtt{Base}\cup\mathtt{Survey}\right)\right)\\
 & \land\boxempty\left(\mathtt{Survey}\rightarrow\ocircle\left(\lnot\mathtt{Survey}\cup\mathtt{Report}\right)\right)\\
 & \land\boxempty\left(\mathtt{Report}\rightarrow\ocircle\left(\lnot\mathtt{Report}\cup\mathtt{Supply}\right)\right).
\end{aligned}
\label{eq:task1}
\end{equation}
In English, $\phi_{s}$ in (\ref{eq:task1}) means the robot needs
to always avoid $\mathtt{Obstacle}$ while repeatedly and sequentially
visiting $\mathtt{Base}$, $\mathtt{Survey}$, $\mathtt{Report}$,
and $\mathtt{Supply}$.

The workspace is abstracted to a grid-like graph consisting of $10\times10$
nodes as shown in Fig. \ref{fig:Snapshots}. The labels $\Pi=\{\mathtt{\mathtt{Base}},\thinspace\mathtt{Supply},\thinspace\mathtt{Report},\thinspace\mathtt{Obstacle},\thinspace\mathtt{Survey}\}$
are shown in circles with blue, purple, cyan, black, and yellow, respectively.
The robot is represented by a red dot transiting along edges between
nodes. Each node $q$ in the graph is associated with a time-varying
reward $R_{k}\left(q\right)$, and the reward is randomly generated
from a uniform distribution in the range $\left[10,25\right]$ at
time $k$. The rewards are presented as green circles with size proportional
to the reward value. LTL2BA \cite{Babiak2012} was used to translate
$\phi_{s}$ to a B\"uchi Automaton $\mathcal{\mathcal{B}}_{s}$ with
$\left|S_{s}\right|=28$ states. 

The simulation was implemented in MATLAB on a PC with 3.6 GHz Quad-core
CPU and 32 GB of RAM. Since the DTS $\mathcal{T}$ has $\left|Q\right|=100$
states, the relaxed product automaton $\mathcal{P}$ has $\left|S_{\mathcal{P}}\right|=2800$
states. The computation of $\mathcal{P}$, the largest self-reachable
set $\mathcal{F}^{*}$, and the energy function took $4.7$s. The
control algorithm outlined in Algorithm \ref{Alg2} was implemented
for 200 time steps with horizon $N=4$. Each iteration of Algorithm
\ref{Alg2} took 1 to 3s depending on the volume of local updates.
To demonstrate the ability of the robot in handling partially infeasible
tasks, it is assumed that the $\phi_{s}$ is fully feasible in the
first 100 time steps and it becomes infeasible afterwards in the sense
that the survey points are not accessible (i.e., yellow nodes are
off).

Fig. \ref{fig:Snapshots}(a) shows the robot's initial knowledge about
the environment, which consists of known destinations $\{\mathtt{Base},\thinspace\mathtt{Supply},\thinspace\mathtt{Report}\}$
and locally observed rewards. Figs. \ref{fig:Snapshots}(b) and (c)
show the snapshots of the environment at $t=1$s and $t=140$s, respectively.
Fig. \ref{fig:Snapshots}(c) shows that $\phi_{s}$ is relaxed since
the robot is required to visit $\mathtt{Survey}$ points in (\ref{eq:task1}),
while $\mathtt{Survey}$ points do not exist from $t=101$s to $t=200$s,
thus leading to a revised motion plan. Note that, due to the consideration
of dynamic obstacles, the deployment of black circles can vary with
time. Figs. \ref{fig:trajectory}(a) and (b) show the trajectories
of the robot in the feasible and infeasible $\phi_{s}$ , respectively.
Fig. \ref{fig:energy} shows the evolution of the energy function
during mission operation. Each time the energy $J\left(s_{\mathcal{P}}\right)=0$
in Fig. \ref{fig:energy} indicates that an accepting state has been
reached, i.e., the desired task is accomplished for one time. The
jumps of energy from $t=100$s to $200$s (e.g., $t=100$s) in Fig.
\ref{fig:energy} are due to the violation of the desired task whenever
the environment becomes infeasible. Nevertheless, the developed control
strategy still guarantees the decrease of energy function to satisfy
the acceptance condition of $\mathcal{P}$. Fig. \ref{fig:reward}
shows the collected local time-varying reward. The simulation video
is provided\footnote{https://www.youtube.com/watch?v=RyRnKXDDH5U\&t=30s}.

\begin{table}
	\textcolor{black}{\caption{\label{tab:scalability}The comparison of workspace size and computation
			time.}
		\centering{}}%
	\scalebox{0.8}{%
	\begin{tabular}{c|cccccc}
		\hline 
		\textcolor{black}{Workspace} & \textcolor{black}{$\mathcal{M}$} & \textcolor{black}{$\mathcal{\mathcal{P}}$} & \textcolor{black}{$Min$} & \textcolor{black}{$Max$} & \textcolor{black}{$Mean$} & \textcolor{black}{Horizon}\tabularnewline
		\textcolor{black}{size{[}cell{]}} & \textcolor{black}{$\left|Q\right|$} & \textcolor{black}{$\left|S_{\mathcal{P}}\right|$} & \textcolor{black}{Time{[}s{]}} & \textcolor{black}{Time{[}s{]}} & \textcolor{black}{Time{[}s{]}} & \textcolor{black}{N}\tabularnewline
		\hline 
		\textcolor{black}{$10\times10$} & \textcolor{black}{100} & \textcolor{black}{2800} & \textcolor{black}{0.88} & \textcolor{black}{2.91} & \textcolor{black}{1.70} & \textcolor{black}{4}\tabularnewline
		\hline 
		\textcolor{black}{$10\times10$} & \textcolor{black}{100} & \textcolor{black}{2800} & \textcolor{black}{1.12} & \textcolor{black}{3.6} & \textcolor{black}{1.81} & \textcolor{black}{6}\tabularnewline
		\textcolor{black}{$30\times30$} & \textcolor{black}{900} & \textcolor{black}{25200} & \textcolor{black}{1.47} & \textcolor{black}{5.45} & \textcolor{black}{3.12} & \textcolor{black}{4}\tabularnewline
		\textcolor{black}{$30\times30$} & \textcolor{black}{900} & \textcolor{black}{25200} & \textcolor{black}{1.99} & \textcolor{black}{9.12} & \textcolor{black}{4.83} & \textcolor{black}{8}\tabularnewline
		\textcolor{black}{$50\times50$} & \textcolor{black}{2500} & \textcolor{black}{700000} & \textcolor{black}{2.01} & \textcolor{black}{14.9} & \textcolor{black}{6.11} & \textcolor{black}{4}\tabularnewline
		\hline 
	\end{tabular}}
\end{table}

\textcolor{black}{In order to demonstrate scalability and computational
	complexity of the framework, we repeat the control synthesis introduced
	above for workspace with different sizes. Specifically, each rectangular
	state of previous $10\times10$ grid-like graph are further divided
	into the number of grid-cells $3^{2}$, $5^{2}$. The dimensions of
	resulting graph, DTS $\mathcal{T}$, and relaxed product automaton
	$\mathcal{P}$, and the maximum, minimum and mean time taken to solve
	the predicted trajectories at each time-step are shown in Table \ref{tab:scalability},
	where we also analysis how the various horizon will influence the
	computations. Due to the off-line computation for largest-self reachable
	sets that always remains the same, we omit its time comparison.}

\textcolor{black}{From table \ref{tab:scalability}, we can see that
	the computational time at each time will be dramatically influenced
	by updated process involving recomputing the energy function based
	on updated knowledge. The benefits of RHC based algorithm in this
	paper is that the proposed algorithm only need to consider the local
	scale optimization problem and the energy constraints will ensure
	the global task satisfaction. As a result, the minimum solving time
	at each time-step will be slightly influenced. }

\subsection{Experiment results\label{subsec:experiment}}

\begin{figure}
\centering{}\includegraphics[scale=0.25]{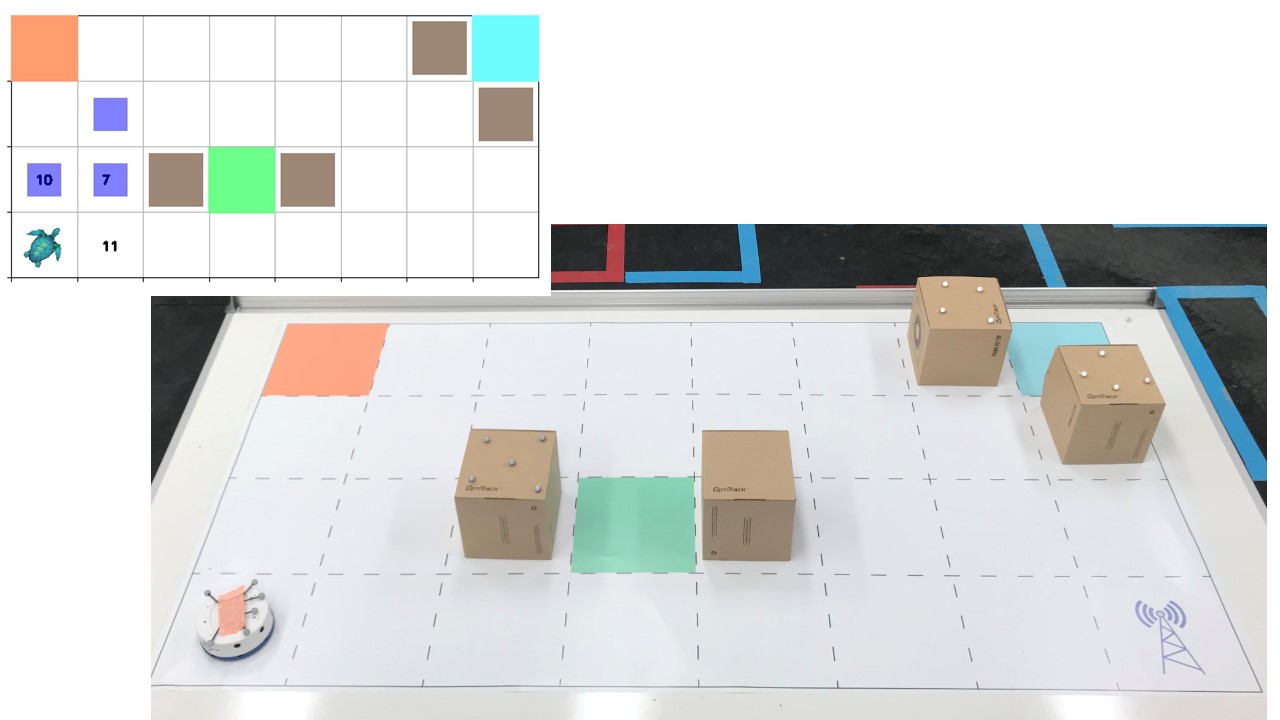}\caption{\label{fig:experiment} The workspace of the real environment (bottom)
and the simulated environment (top). The turtle represents the robot,
and the blue squares represent the predicted trajectory, where the
number indicates the locally observed rewards.}
\end{figure}

Experiments were performed on a mobile robot, Khepera IV, to verify
the developed control strategy. The workspace is about $48''\times96''$,
consisting of $4\times8$ square cells, as shown in Fig. \ref{fig:experiment},
where the bottom figure shows the experiment workspace while the top
figure shows the corresponding simulated workspace. The robot is allowed
to transit between adjacent cells, i.e., the robot at a cell has four
possible actions, ``up,'' ``down,'' ``right,'' and ``left.''
Consider three areas of interest, $P_{1}$, $P_{2}$, and $P_{3}$,
which correspond to orange, green, and cyan cells, respectively. The
desired task of the robot is to avoid obstacles (i.e., carbon boxes
in Fig. \ref{fig:experiment}) and visit $P_{1}$, $P_{2}$, and $P_{3}$
sequentially and infinitely often, which is expressed as an LTL specification
\begin{equation}
\begin{aligned}\phi_{s}= & \boxempty\diamondsuit P_{1}\\
 & \land\boxempty P_{1}\rightarrow\ocircle\left(\lnot P_{1}\cup P_{2}\right)\\
 & \land\boxempty P_{2}\rightarrow\ocircle\left(\lnot P_{2}\cup P_{3}\right).
\end{aligned}
\label{eq:task2}
\end{equation}
The robot is assumed to know the locations of $P_{1}$, $P_{2}$,
and $P_{3}$, without knowing the obstacle positions. It is possible
that the preassigned task $\phi$ cannot be fully accomplished, due
to unexpected obstacles. Fig. \ref{fig:experiment} shows an infeasible
case of $\phi_{s}$, where $P_{3}$ is surrounded by obstacles and
not accessible by the robot. Therefore, the task (\ref{eq:task2})
cannot be fully realized, and the robot has to revise its motion plan
and adapts to the real environment. In addition, each cell is assumed
to have a time-varying reward, randomly generated from a uniform distribution
in the range $\left[5,15\right]$. The robot is desired to maximize
reward collection while performing the task (\ref{eq:task2}).

The online motion planning strategy in Algorithm \ref{Alg2} was implemented
in python on a VMware with a 3.6 GHz Quad-core CPU and 8 GB of RAM.
The robot actuation module was implemented on Linux with an Optitrack
motion capture system providing real-time position feedback of the
robot. The B\"uchi Automaton $\mathcal{\mathcal{B}}_{s}$ has $\left|S_{s}\right|=12$
states, and the DTS $\mathcal{T}$ has $\left|Q\right|=32$. The relaxed
product automaton $\mathcal{P}$ has $\left|S_{\mathcal{P}}\right|=384$
states. The horizon in RHC was selected as $N=4$, and the computation
of Algorithm \ref{Alg2} at each iteration took $0.25$s. During implementation
of Algorithm \ref{Alg2}, the obstacles can be randomly moved, and
the robot usually took about $0.5$s to update its motion plan. The
experiment video is provided\footnote{https://youtu.be/16j6TmVUrTk}.

\section{Conclusions}

An RHC-based online motion planning strategy with partially infeasible
LTL specifications is developed in this work to enable the autonomous
robot to maximize reward collection while considering hard and soft
LTL constraints. Motion planning in an uncertain environment can be
better modeled by a Markov decision process. Future research will
consider extending this work with more realistic robot models and
advanced learning based motion planning . Additional research will
also consider extending the current work to continuous state space
using hybrid control.

\subsection*{Acknowledgments}

We thank Marius Kloeatzer, Xuchu Ding, and Calin Belta for their software
and open source. 

\bibliographystyle{IEEEtran}
\bibliography{BibMaster}

\end{document}